%% file: _main.tex
\documentclass[11pt]{article}
\usepackage{acl}
\usepackage{times}
\usepackage{todonotes}
\usepackage{pdfcomment}
\usepackage{latexsym}
\usepackage{caption}
\usepackage{subcaption}

\usepackage[T1]{fontenc}
\usepackage[utf8]{inputenc}
\usepackage{microtype}
\usepackage{inconsolata}
\usepackage{graphicx}
\usepackage{booktabs}
\usepackage{longtable}
\usepackage{array}
\usepackage{tcolorbox}

\title{SafeTalkCoach: Diversity-Driven Multi-Agent Simulation for \\Parent-Teen Health Conversations}

\author{
\textbf{Benyamin Tabarsi\textsuperscript{1*}, Wenbo Li\textsuperscript{1*}, Tahreem Yasir\textsuperscript{1}, Aryan Santhosh Kumar\textsuperscript{1}} \\
\textbf{Laura Widman\textsuperscript{2}, Dongkuan (DK) Xu\textsuperscript{1}, Tiffany Barnes\textsuperscript{1}} \\
\textsuperscript{1}North Carolina State University \\
\textsuperscript{2}Florida State University \\
\texttt{\{btaghiz,wli55,tyasir,asantho,dxu27,tmbarnes\}@ncsu.edu} \\
\texttt{lwidman@fsu.edu} \\
\small \textsuperscript{*}Both authors contributed equally to this research.
}

\begin{document}
\maketitle
\input{sections/0_abstract}
\input{sections/1_introduction}

\input{sections/2_related_work}
\input{sections/3_methodology}
\input{sections/4_experimental_results}
\input{sections/5_conclusion}
\input{sections/limitations}

\input{sections/ethical_considerations}
\bibliographystyle{acl_natbib}
\bibliography{sections/references, software}
\appendix
\input{sections/ablation_setup}

\input{sections/engagement_score}
\input{sections/data_collection}
\input{sections/baseline_setup}
\input{sections/human_annotation_process}
\input{sections/dataset_statistics}

\input{sections/dialogue_examples}

\input{sections/coding_instruction}
\input{sections/prompt_examples}

%%
%% If your work has an appendix, this is the place tow   put it.

\end{document}

%% file: sections/0_abstract.tex
\begin{abstract}
The importance of effective parent-child communication about sexual health is widely acknowledged, but real-world data on these conversations is scarce and challenging to collect, due to their private and sensitive nature. Although LLMs have been widely adopted in dialogue generation, they may deviate from best practices and frequently lack realism and diversity. We introduce SafeTalkCoach, a diversity-driven multi-agent dialogue generation framework that simulates parent-child conversations about sexual health, and present an accompanying dataset. SafeTalkCoach integrates crowd-sourced and synthesized scenarios, established sexual health guidelines, evidence-based personas, adaptive control modules, and hierarchical diversification. Through evaluations, we demonstrate that SafeTalkCoach generates diverse conversations while maintaining realism, communication quality, and controllability in practice. Our goal is that the SafeTalkCoach framework and the dataset support both AI research and health communications practices. 
\end{abstract}

%% file: sections/1_introduction.tex
\section{Introduction} 
\begin{figure}[t!]
    \centering
    \includegraphics[width=\linewidth]{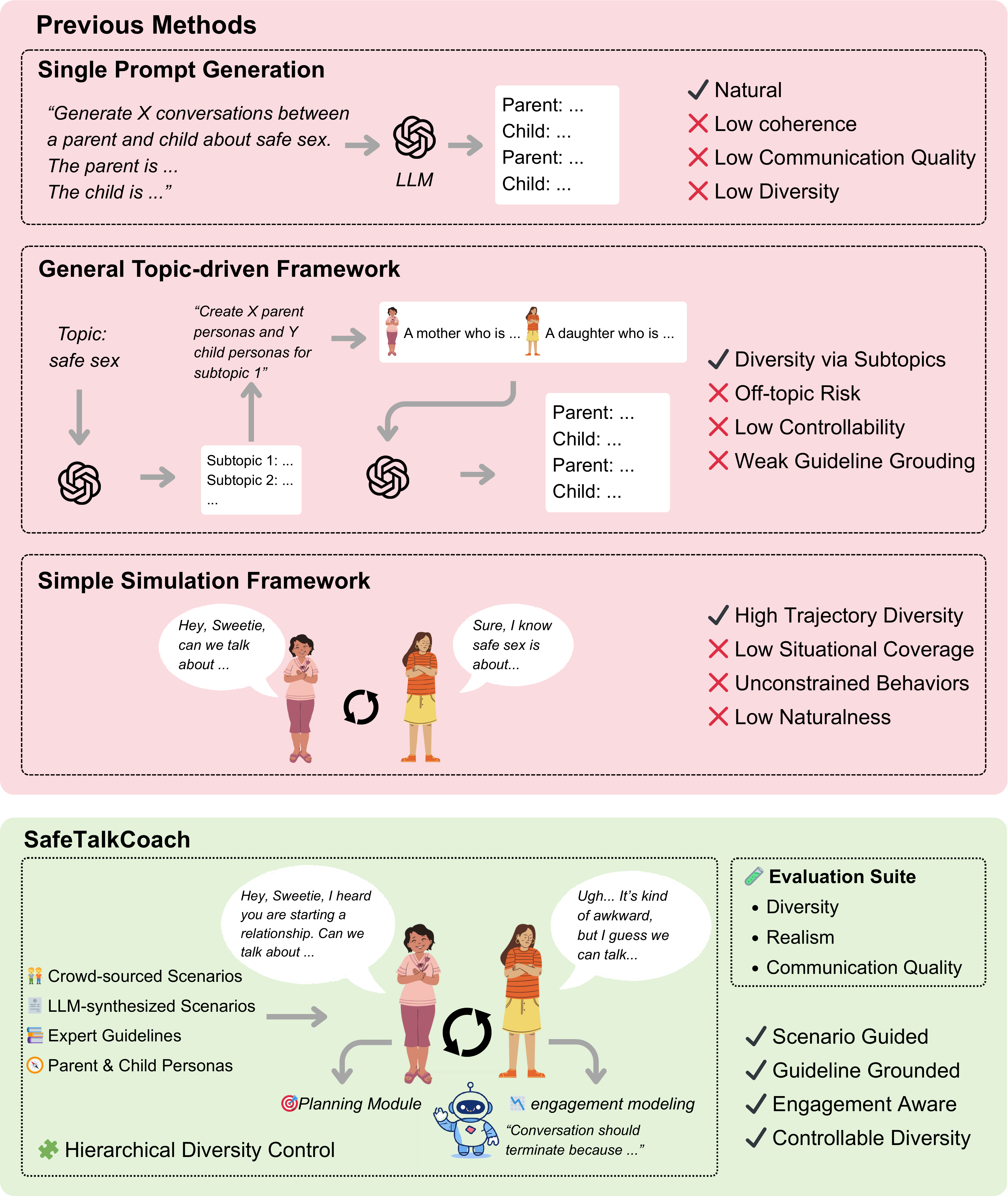}
    \caption{SafeTalkCoach vs. Existing Frameworks}
    \label{fig:compairson}
\end{figure}

Effective, open parent–child conversations about sexual health offer numerous benefits for adolescents' sexual knowledge and well-being \cite{widman_assessment_2019,rogers_parentadolescent_2017,d_parent-based_2015}. Research indicates that such conversations are associated with increased contraceptive use \cite{rogers_parentadolescent_2017,harris_parental_2013}, more awareness of the consequences of sexual activity \cite{rogers_parentadolescent_2017}, and enhanced willingness to discuss sexual issues with parents \cite{isaksen_parent-child_2020}. However, progress in understanding and improving these conversations is constrained by a data bottleneck. Collecting authentic conversational data faces severe barriers such as privacy concerns and discomfort with on-record discussions of intimate topics \cite{d_parent-based_2015,mullis_international_2021}. This scarcity of data prevents researchers from identifying effective communication strategies, developing automated interventions, or training counselors on realistic scenarios. 

Developments in large language models (LLMs) over recent years have presented new avenues for synthesizing human-like conversations. Nonetheless, several issues undermine their utility for research and training. First, Conversations generated by LLMs may deviate from evidence-based guidance, 
as exemplified by a large-scale mental health QA benchmark identifying their overgeneralization and speculative diagnoses \cite{li_counselbench_2025}. Second, LLMs struggle with conversational realism and naturalness, such as generating short and more predictable utterances in Prolonged Exposure (PE) therapy dialogues \cite{bn_how_2025}. Finally, conversations often converge on prototypical scripts, as observed in prior research \cite{chandra_reasoning_2025}, where even state-of-the-art reasoning LLMs have been shown to be limited in adapting to diverse patient personas and needs in multi-turn mental health conversations. 

These shortcomings prevent synthetic dialogues from representing the behavioral and topical spectrum that practitioners most need to study. However, addressing these limitations is non-trivial. First, the black-box nature of LLMs makes it challenging to ensure adherence to best practices in high-stakes fields, such as those involving mental health \cite{sun_script-strategy_2025}. Second, realism in LLM outputs requires dynamic and context-aware modeling rather than static prompts \cite{lu_can_2025}. Third, meaningful diversity requires controlling context and model behavior, not just varying content \cite{liu_controllable_2024}.

In this paper, we propose SafeTalkCoach, a multi-agent framework that generates diverse, realistic, and guideline-adherent parent-child dialogues about sexual health, and release a dataset created with it. To address the lack of grounded behaviors and realism, we decompose roles into three LLM-powered agents with distinct responsibilities and control modules: (1) a Parent Agent that starts and maintains conversations about specific scenarios regarding sexual health, following evidence-based sexual health communication guidelines; (2) a Child Agent that responds with adaptable and stochastic engagement states; and (3) a Moderator Agent that monitors the conversation and terminates it when appropriate. To address the homogeneity problem, we propose a hierarchical diversification mechanism that operates at three levels: context-level diversity through synthetic and crowd-sourced scenarios, behavioral-level diversity through agent control modules, and response-level diversity through integration of verbalized sampling \cite{zhang_verbalized_2025}. 

Through our evaluation, we demonstrate that SafeTalkCoach's use of hierarchical diversification and agent-level control modules supports controllable, focused, and high-quality simulations across a wide range of scenarios, addressing challenges observed in other frameworks. Our primary contributions can be summarized as follows:
\begin{itemize}
    \item We design and implement SafeTalkCoach, a multi-agent simulation framework with agent-level control and hierarchical diversification for parent-child conversations about sexual health.
    % framework + advantage
    \item We present a dataset of 1,495 SafeTalkCoach-generated parent-child conversations covering three representative discussion topics. 
    % dataset
    \item We conduct comprehensive evaluations demonstrating that SafeTalkCoach generates more diverse, realistic, and guideline-adherent dialogues compared to baseline approaches, while avoiding topic drift and low controllability.
\end{itemize}

%% file: sections/2_related_work.tex
\section{Related Work}
\label{related_work}
\subsection{Parent-Teen Health Communication and Data Challenges}
Parent-child communication about sexual health is often hindered by embarrassment, discomfort, and cultural taboos \cite{mullis_international_2021}. This sensitive nature of parent–teen sexual health communication makes collecting such data challenging for researchers \cite{d_parent-based_2015}. With privacy considerations, prior works have collected human dialogue data in related domains, such as SpecialTime for parent-child therapy conversations \cite{huber_specialtime_2019} and SpokenWOZ for spoken, human-to-human task-oriented dialogues \cite{si_spokenwoz_2023}. However, these approaches are not focused on sexual health conversations, where privacy concerns and sensitivities are even more pronounced. These challenges necessitate careful data synthesis approaches that model realistic and effective parent-child interactions.

% and difficulties in capturing authentic interpersonal dynamics 
\subsection{LLM-Based Dialogue Generation Frameworks}
\label{related_work_frameworks}
LLMs have been extensively used by researchers to generate dialogue datasets. The focus of the literature ranges from dialogue reconstruction based on psychological counseling reports \cite{zhang_cpsycoun_2024} and doctor-patient dialogue synthesis based on clinical notes \cite{wang_notechat_2024}, to synthesizing therapist-client dialogues based on Motivational Interviewing theory \cite{kim_kmi_2025}. However, these approaches rely on structured, domain-specific reference data and are therefore difficult to adapt to other domains.

With a domain-agnostic lens, DiaSynth \cite{suresh_diasynth_2025} introduces a dialogue generation framework that creates subtopics and corresponding personas, followed by dialogue synthesis via the Chain-of-Thought (CoT) approach. However, their approach cannot model the nuances of sensitive parent–teen conversations, such as the child's engagement and the parent's attitude. 

We address these limitations by grounding dialogue generation in authentic, comprehensive contexts to mitigate the lack of reference data. By integrating evidence-based guidelines, research-backed personas, and adaptive control modules, our framework ensures alignment with evidence-based best practices while maintaining dynamic and realistic behaviors, enabling the utility of generated dialogues by practitioners.

\subsection{Diverse Dialogue Generation}
Lack of diversity in LLM responses has been widely identified, largely due to \textit{Mode collapse} \cite{janus_mysteries_2022,omahony_attributing_2024}, where a model concentrates on a few high-probability responses.

Zhang et al. \cite{zhang_verbalized_2025} addressed this with \textit{Verbalized Sampling}, prompting LLMs to produce multiple scored responses and thereby improving output diversity across tasks, including dialogue simulation. However, these works focus primarily on response-level lexical and content diversity, which is insufficient for parent-child conversation datasets to be useful. 

Our work extends diversity from the response level to the dialogue level, using a hierarchical diversification pipeline to generate dialogues that encompass various scenarios, agent behaviors, and utterances, providing a wide contextual and behavioral spectrum for practitioners to study.

%% file: sections/3_methodology.tex
\section{The Proposed Method: SafeTalkCoach}
We introduce SafeTalkCoach, a persona-based multi-agent framework for generating realistic parent-child dialogues grounded in established sexual health communication guidelines. The framework comprises three agents, i.e., a parent, a child, and a moderator. To reflect the heterogeneity of real-world interactions, we integrate hierarchical diversification at the data, architecture, and prompting levels throughout the pipeline.

\begin{figure*}
    \centering
    \includegraphics[width=1.0\linewidth]{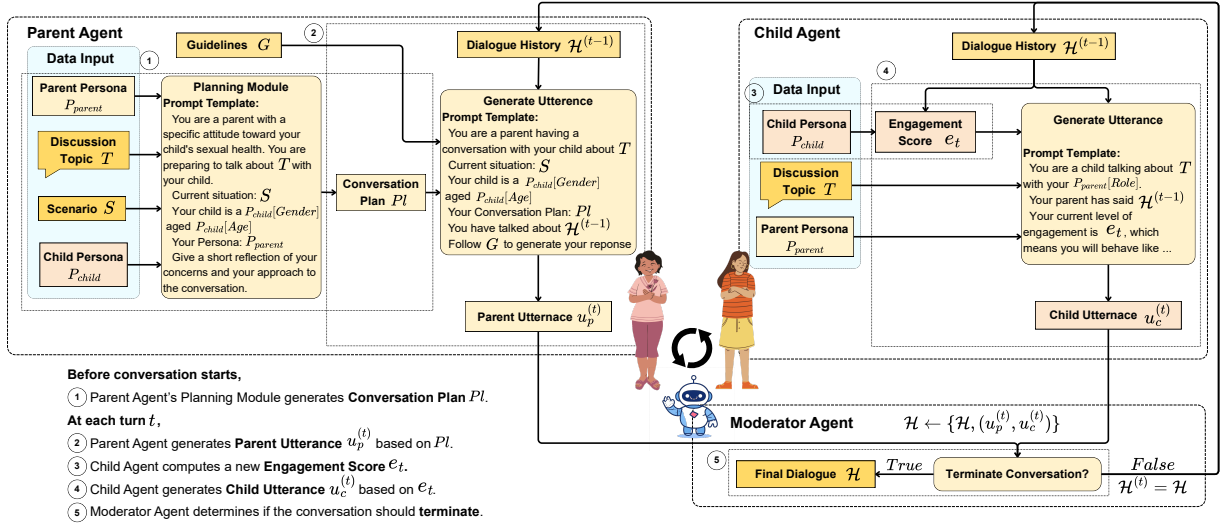}
    \caption{SafeTalkCoach's Multi-Agent Dialogue Generation Pipeline}
    \label{fig:architecture}
\end{figure*}

\subsection{Multi-Agent Dialogue Generation Pipeline}
SafeTalkCoach generates a dialogue turn-by-turn. The parent agent initiates and steers the conversation toward intended discussion points; the child agent responds with age- and attitude-appropriate language controlled by a dynamic engagement state; and the moderator agent terminates the dialogue once it reaches a natural endpoint.

\subsubsection{Parent Agent}
We model the parent agent along two persona dimensions that research \cite{scull_exploring_2022,astle_parent-child_2023} has shown to influence sexual health conversations: role (\textit{mother} vs. \textit{father}) and general attitude toward sexuality (\textit{traditional} vs. \textit{progressive}). Each dialogue begins with a scenario that specifies the triggering event and the discussion points the parent intends to address.

\paragraph{Planning Module.} Before the dialogue begins, a Planning Module takes the parent persona, scenario, and discussion topic as input and uses an LLM to generate a conversation plan. The plan functions as a scenario-conditioned persona, which provides a high-level roadmap and maintains the agent’s behavior consistency across turns. 

At each turn, the parent generates an utterance conditioned on the conversation plan, dialogue history, and the child’s age and gender. To ground responses in best practices, we instruct the agent using guidelines from Planned Parenthood \cite{plannedparenthood_tips_2025} and Crucial Conversations \cite{chepul_crucial_2021}.

\subsubsection{Child Agent} 
The child agent is parameterized by gender (\textit{male} or \textit{female}), age (11–18, grouped as \textit{11–13} or \textit{14–18}), and baseline attitude toward sexual health conversations (\textit{dismissive/resistant} vs. \textit{somewhat comfortable}). Prior works~\cite{flores_21st_2017, malacane_review_2016,wilson_mothers_2010} identify these dimensions as key factors influencing adolescent responses in sensitive discussions.

\paragraph{Engagement Score Module.} Engagement in a conversation can be influenced by personality, 
the dynamics of the conversation, and exogenous factors unrelated to the conversation \cite{salam_automatic_2022, inoue_engagement_2018}. To capture this, we introduce a heuristic-based Engagement Score Module that maintains a latent engagement score and uses it to modulate the child’s responsiveness. 

\begin{figure}[t!]
    \centering
    \includegraphics[width=1.0\linewidth]{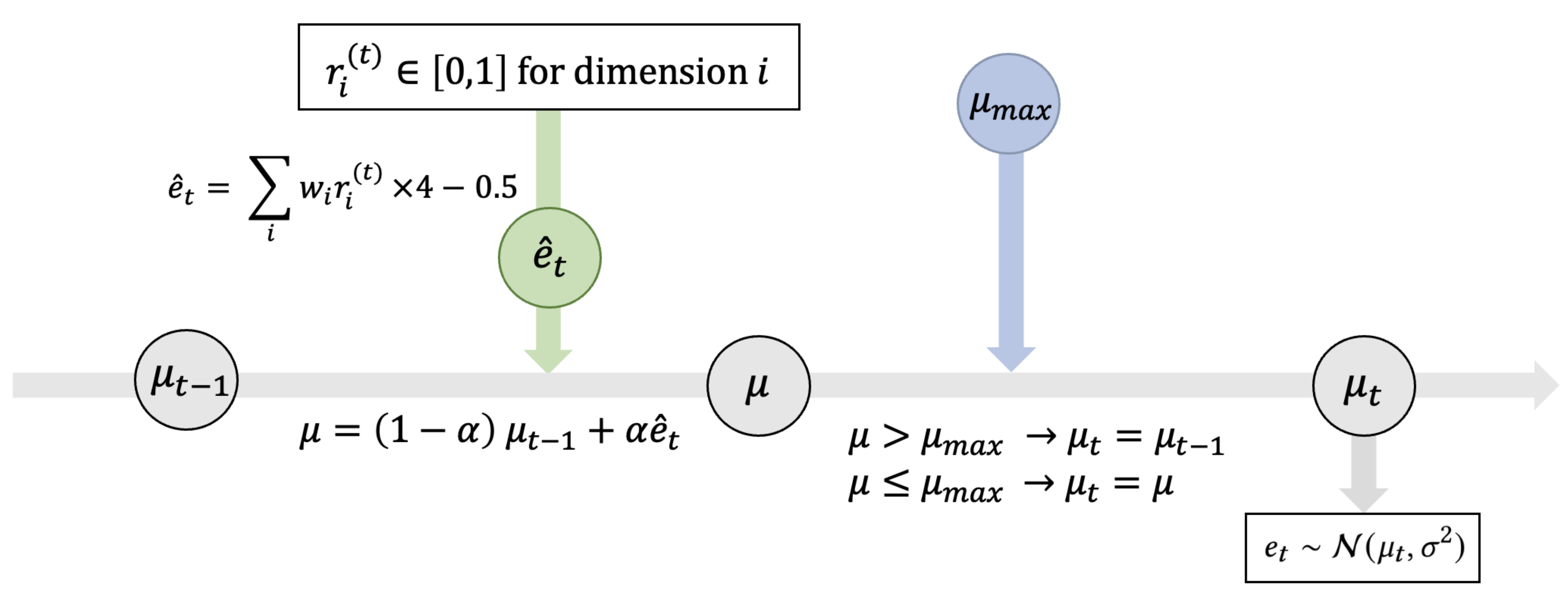}
    \caption{Heuristic-based Engagement Score Module}
    \label{fig:engagement}
\end{figure}

As shown in Figure \ref{fig:engagement}, at each turn $t$, the engagement score $e_t$ is sampled from a Gaussian distribution, which supports continuous and smooth latent representation of engagement with stochasticity and avoids frequent jumps. The mean of the distribution $\mu_t$ represents the child’s most probable internal engagement state, and the variance $\sigma^2$ controls the stochasticity of engagement across turns. 

Before sampling at each turn $t$, the child agent first rates the parent’s utterance along each engagement-relevant dimension $i$ informed by prior research and guidance \cite{guilamo-ramos_parent-adolescent_2008, maina_parent-child_2020, plannedparenthood_tips_2025}, producing a set of ratings $r_i^{(t)}$. The ratings are then transformed into an estimate $\hat{e}_t$, interpreted as the most probable engagement score after hearing the parent’s utterance under ideal conditions. Finally, we update the mean to $\mu_t$ via a smooth linear rule using $\hat{e}_t$ to avoid abrupt state changes.

To further encode the diversity of children's responsiveness and openness, we set a random dialogue-specific maximum mean $\mu_{max}$, which stops the increment of $\mu_{t}$ if it would be pushed above $\mu_{max}$. Details are provided in Appendix \ref{appx:engagement}.

At each turn, the child agent is provided with the discussion topic, their parent's role, dialogue history, and their current engagement score to generate the next utterance.

\subsubsection{Moderator Agent}
We design a moderator agent to determine when the conversation should terminate based on either of two criteria: 1) consecutive disengagement of the child, or 2) the child reflects on the key points the parent has covered with no further questions.

\subsection{Hierarchical Diversification Mechanism}
We employ a hierarchical diversification mechanism to ensure that the generated dialogues encompass a wide variety of scenarios, personas, and behaviors. The mechanism has 3 levels:

\paragraph{Data Level.} 
Strong situational coverage is essential for guiding parents effectively in various real-life circumstances. To diversify the conversation contexts, we collect scenarios and personas from a combination of real-world sources (for authenticity and ecological validity) and research-based synthesis (for systematic coverage).

\paragraph{Architectural Level.}
Parents differ in their communication styles and goals, and children vary in their openness to engagement, resulting in diverse behaviors and interaction patterns. The control modules in our agents, i.e., the planning module and the engagement score module, capture such diversity by generating dialogue-specific plans and simulating dynamic behaviors.

\paragraph{Prompting Level.}
We implement scenario synthesis and the Planning Module with Verbalized Sampling to reduce mode collapse and generate multiple 
% coherent 
responses from the same specification.

\subsection{Hybrid Data Collection Pipeline}
We focus on three representative topics commonly recommended by sexual health education curricula \cite{cdc_hecat_2021, fose_national_2020}: \textit{safe sex}, \textit{abstinence}, and \textit{consent}. To our knowledge, there is no reliable data source that directly provides scenarios, personas, or model conversations. To overcome this limitation, we collect scenarios following a hybrid approach, where scenarios and personas are both synthesized using categories defined in research to ensure comprehensiveness and extracted from real-world crowd-sourced data to improve authenticity. 

\paragraph{LLM-Synthesized Data.} We base our scenario synthesis on Teachable Moments from health behavioral change theory \cite{lawson_teachable_2009}. We follow four broad categories of teachable moments in sexual health (\textit{Major Life Events}, \textit{Social Media}, \textit{Mass Media}, and \textit{Everyday Occurrences}) as suggested by prior research \cite{ashcraft_talking_2017}. We instruct OpenAI's GPT-4o mini model \cite{openai_gpt-4o-mini_2025} (with temperature 0.8 for maintaining format consistency in long outputs) to synthesize scenarios from each category for each topic, providing examples from the literature. This yields 425 scenarios covering various situations parents might face. 
\paragraph{Crowd-sourced Data.} We extract posts from Reddit \cite{reddit_reddit_2025} and StackExchange \cite{stackexchange_parenting_2025} as platforms where parents discuss their real concerns.

To filter out only high-quality posts, an LLM performs an initial screening, followed by two authors manually annotating the extracted posts on their relevance to the topics, parent and child personas, and the corresponding teachable moments, which were then fed into an LLM to generate scenario descriptions. This process yields 74 highly authentic scenarios. More details could be found in Appendix \ref{appx:collection}.

\begin{figure}[t]
    \centering
    \includegraphics[width=\linewidth]{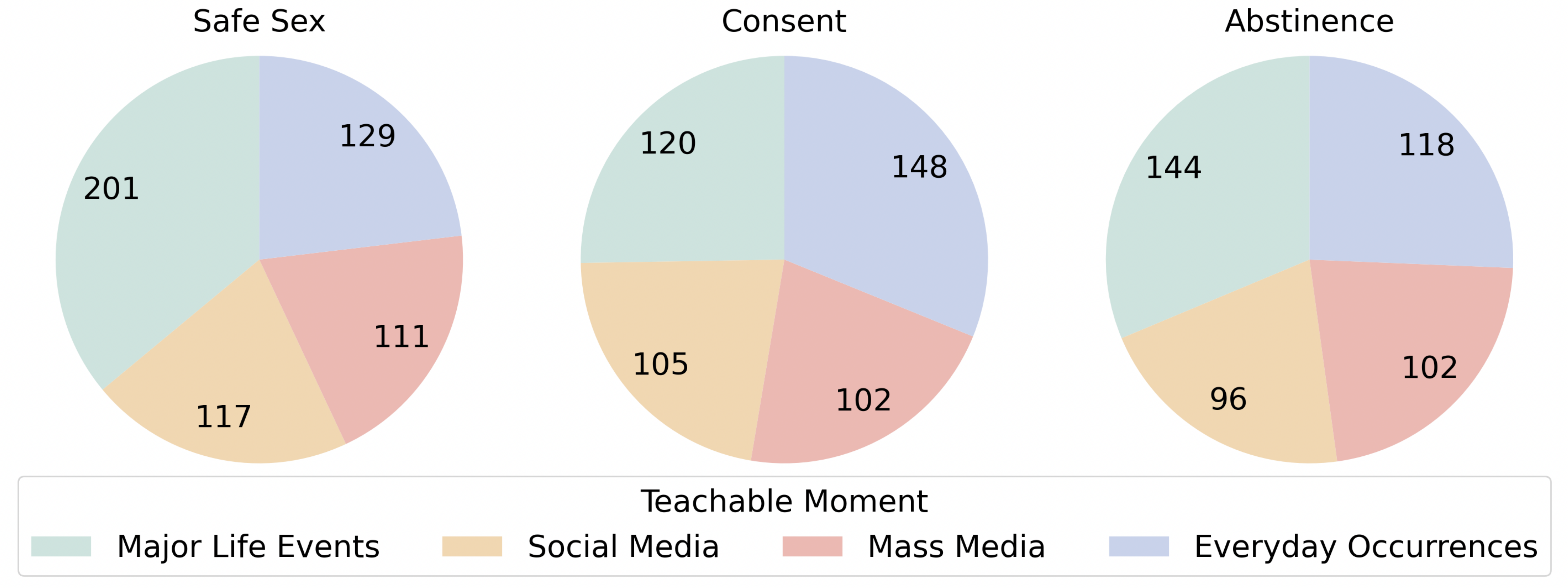}
    \caption{Distribution of our Dataset}
    \label{fig:distribution}
\end{figure}

Each scenario was fed into GPT-4o mini (with temperature 1 for maximizing randomness) to generate three dialogues, resulting in a dataset of 1,495 dialogues. As shown in Figure \ref{fig:distribution}, the dialogues encompass a wide range of teachable moments in each topic. Dataset statistics and example dialogues are presented in Tables \ref{tab:dataset_statistics} and \ref{tab:safetalkcoach_examples}, respectively.

%% file: sections/4_experimental_results.tex
\section{Experiments and Results}
\label{experimental_results}
Evaluating parent-child conversations on sensitive topics presents unique challenges, since no human-annotated ground truth datasets exist, benchmarks are scarce, and evaluation dimensions are often vague and subjective. Therefore, we combine large-scale automated and LLM-based assessment, supplemented by human evaluation of common communication mistakes on a subset of data.
All LLM-based evaluations are conducted using OpenAI's GPT-5 mini model \cite{openai_gpt-5-mini_2025}.

\subsection{Baselines}
To enable a meaningful evaluation in the absence of established frameworks for parent–child conversations on sensitive topics, we construct a set of baselines that covers both: (1) simple domain-specific dialogue generation strategies as lower bounds, and (2) domain-agnostic dialogue generation frameworks with minimal modifications to fit into our task. Detailed setup is discussed in Appendix \ref{appx:baseline}. Specifically, we have these baselines:

\begin{itemize}
    \item \textbf{Direct Prompting} \textit{(Direct)}. One LLM generates multiple dialogues in a single prompt. 
    \item \textbf{Procedural Prompting} \textit{(Procedual)}. One LLM generates multiple conversation starters, followed by separate prompts to expand them. 
    \item \textbf{PSYDIAL} \cite{han_psydial_2024}, the first Korean personality-based dialogue generation framework that  focuses on the Extraversion dimension of the Big Five personality traits.
    \item \textbf{DiaSynth} \cite{suresh_diasynth_2025} which diversifies dialogues by creating a few subtopics and corresponding personas before generation.
    \item \textbf{Family Conversation Simulation} \textit{(FCS)} \cite{ye_simulating_2024}, a multi-agent simulation framework that demonstrates family conversations with minimum control mechanisms. 
\end{itemize}

% For each baseline, we generate and sample an average of 250 dialogues per topic using GPT-4o-mini. More details are in Appendix \ref{appx:baseline}

\subsection{Automated Evaluation}
We evaluate three dimensions of generated dialogues: \textit{diversity}, \textit{realism}, and \textit{communication quality} using automated evaluation metrics.

\subsubsection{Evaluation Metrics}
\label{experiment_setup}

\paragraph{\textbf{Diversity}} Prior works assessing diversity evaluate it at the response level with metrics such as pairwise cosine similarity and self-BLEU \cite{tevet_evaluating_2021, zhang_evaluating_2025, zhang_verbalized_2025}. Hence, we evaluate response-level diversity by embedding the dialogues using OpenAI's \textit{text-embedding-3-small} model \cite{openai_text-embedding-3-small_2025}, and computing $1-$pairwise cosine similarity of the embeddings between dialogues from the same approach.

Embedding similarity captures only surface-level variation, not whether dialogues incorporate diverse real-world conversation contexts, interaction patterns, and personas. We therefore complement our analysis with dialogue-level diversity using an LLM-based assessment. Built upon prior work \cite{zhang_evaluating_2025}, we implement an LLM-based leaderboard to perform pairwise comparison between our data and data from baselines, focusing on three dimensions: \textbf{Situational Coverage} \textit{(Situational)}, \textbf{Parental Strategy Diversity} \textit{(Strategy)}, and \textbf{Dialogue Trajectory Diversity} \textit{(Trajectory)}. 

% \begin{itemize}
%     \item \textbf{Situational Coverage} \textit{(Situational)}: diversity of the context, trigger, stakes, and goals of the dialogues.
%     \item  \textbf{Parental Strategy Diversity} \textit{(Strategy)}: distinct communication strategies adopted by the parent.
%     \item \textbf{Dialogue Trajectory Diversity} \textit{(Trajectory)}: diversity of the overall flow and outcomes of the dialogues.
% \end{itemize}

To mitigate ordering bias, we instruct the LLM to provide a score from 1 to 5 on each dimension and overall diversity for each 
set of dialogues rather than a ranking, and randomly swap the two sets.

\paragraph{\textbf{Realism}} 
We evaluate realism using the reference-free LLM evaluation framework G-Eval \cite{liu_g-eval_2023} adopted by prior work \cite{suresh_diasynth_2025} for assessing dialogue quality. We assess two dimensions in G-Eval: \textbf{naturalness} \textit{(Nat.)} and \textbf{coherence} \textit{(Coh.)}, and instruct the LLM to score each dialogue from 1 to 3 on each dimension, reporting the mean score for dialogues from each approach.

We also evaluate local consistency using an entailment-based metric. For each adjacent utterance pair, we apply DeBERTa-v3-base-mnli \cite{moritz_deberta-v3-base-mnli_2025} and compute the rate of contradictions (Contr.) for dialogues from each approach. This captures local logic comparability, complementing G-Eval that assesses global coherence.

% For dialogue datasets to be useful, they must be realistic, coherent, and natural. 

\paragraph{\textbf{Communication Quality}}

Finally, we assess whether parents in the dialogues follow evidence-based best practices. Drawing on previous research \cite{javidi_parent-teen_2025}, we evaluate the communication quality of the dialogues, in particular, the parents' behaviors, on whether they deliver \textbf{caring and warmth} to their children \textit{(Warmth)}, convey \textbf{helpful information} \textit{(Info.)}, and \textbf{Encourage their children's engagement} \textit{(Engage)}.

% Assessing whether dialogues exhibit appropriate communicative behaviors is necessary to ensure alignment with best practices and usability of the dataset by practitioners. 

% \begin{itemize}
%     \item Deliver \textbf{caring and warmth} to their children \textit{(Warmth)}.
%     \item Convey \textbf{helpful information} \textit{(Info.)}.
%     \item \textbf{Encourage their children's engagement} in the conversation \textit{(Engage)}.
% \end{itemize}

Given the customized nature of these dimensions, we design our own evaluation instructions with a similar structure to the G-Eval framework. We instruct the LLM to provide a score from 1 to 5 on each dimension for a dialogue, reporting the mean score for dialogues from each approach.

% \subsubsection{Baselines}

% \subsubsection{Results}

\subsubsection{Results}
\paragraph{Diversity} On embedding-based diversity (Table \ref{tab:embedding_diversity}), Direct Prompting scores highest at 0.4099, while PSYDIAL produces highly clustered outputs, resulting in a low diversity score of 0.1766, which demonstrates a classic mode collapse issue. Other frameworks fall between these extremes (0.23–0.30), with SafeTalkCoach at 0.30.

\begin{table}[t]
\centering
\footnotesize
\setlength{\tabcolsep}{6pt}
\renewcommand{\arraystretch}{1.15}
\begin{tabular}{lc}
\toprule
\textbf{Framework} & \textbf{Embedding Diversity} \\
\midrule
STC & 0.3033 \\
Direct & \textbf{0.4099} \\
Procedural & 0.2981 \\
PSYDIAL & 0.1766 \\
DiaSynth & 0.2983 \\
FCS & 0.2352 \\
\bottomrule
\end{tabular}
\caption{Embedding-based diversity scores.}
\label{tab:embedding_diversity}
\end{table}

\begin{table*}[t]
\centering
\footnotesize
\setlength{\tabcolsep}{4pt}
\renewcommand{\arraystretch}{1.15}
\resizebox{\textwidth}{!}{
\begin{tabular}{lcccccccccccc}
\toprule

& \multicolumn{4}{c}{STC vs Direct}
& \multicolumn{4}{c}{STC vs Procedural}
& \multicolumn{4}{c}{STC vs PSYDIAL} \\
\cmidrule(lr){2-5}
\cmidrule(lr){6-9}
\cmidrule(lr){10-13}

Dimension
& STC-W & BL-W & STC-A & BL-A
& STC-W & BL-W & STC-A & BL-A
& STC-W & BL-W & STC-A & BL-A \\
\midrule

Overall
& 100\%  & 0.0\%  & 4.67 & 2.47
& 86.7\% & 6.7\%  & 4.40 & 3.07
& 93.3\% & 0.0\%  & 4.53 & 2.73 \\

Situational
& 100\%  & 0.0\%  & 4.87 & 2.87
& 93.3\% & 0.0\%  & 4.80 & 3.13
& 93.3\% & 0.0\%  & 4.80 & 2.93 \\

Strategy
& 100\%  & 0.0\%  & 4.67 & 2.40
& 80.0\% & 6.7\%  & 4.40 & 3.27
& 93.3\% & 0.0\%  & 4.53 & 2.80 \\

Trajectory
& 100\%  & 0.0\%  & 4.67 & 2.33
& 73.3\% & 6.7\%  & 4.00 & 3.00
& 86.7\% & 6.7\%  & 4.20 & 2.67 \\

\midrule

\multicolumn{13}{c}{
\begin{tabular}{lcccccccc}
\toprule
& \multicolumn{4}{c}{STC vs DiaSynth}
& \multicolumn{4}{c}{STC vs FCS} \\
\cmidrule(lr){2-5}
\cmidrule(lr){6-9}

Dimension
& STC-W & BL-W & STC-A & BL-A
& STC-W & BL-W & STC-A & BL-A \\
\midrule

Overall
& 33.3\% & 33.3\% & 3.93 & 4.00
& 26.7\% & 60.0\% & 3.67 & 4.13 \\

Situational
& 53.3\% & 46.7\% & 4.27 & 4.20
& 33.3\% & 20.0\% & 4.47 & 4.20 \\

Strategy
& 33.3\% & 53.3\% & 3.80 & 3.93
& 26.7\% & 73.3\% & 3.73 & 4.47 \\

Trajectory
& 40.0\% & 46.7\% & 3.60 & 3.73
& 13.3\% & 80.0\% & 3.00 & 4.07 \\

\bottomrule
\end{tabular}
} \\

\bottomrule
\end{tabular}
} 
\caption{LLM-based diversity comparison between \textbf{SafeTalkCoach (\textit{STC})} and \textbf{Baseline (\textit{BL})} generation frameworks.
\textit{STC-W / BL-W} denote \textbf{win rates}, and \textit{STC-A / BL-A} denote \textbf{mean diversity scores}}
\label{tab:diversity-two-part}
\end{table*}

However, embedding diversity reflects surface-level content variation, not structural and interaction patterns. For example, despite direct prompting achieving the highest level of embedding diversity, the dialogues differ superficially on discussion points with little context and follow nearly identical interaction patterns, as shown in Table \ref{tab:direct_examples}.
% In our observations, we found that embedding-based metrics capture topical breadth but overlook conversation structure and interaction patterns entirely, which we argue are more valuable for practitioner training and downstream tasks, such as model fine-tuning. 

On dialogue-level diversity (Table \ref{tab:diversity-two-part}), SafeTalkCoach outperforms Direct, Procedural, and PSYDIAL, winning 87–100\% of pairwise comparisons. Mean scores confirm the pattern, with SafeTalkCoach scoring 4.4–4.7 versus 2.5–3.1 for baselines.

SafeTalkCoach and DiaSynth achieve comparable overall diversity (3.93 vs. 4.00), but through different mechanisms. SafeTalkCoach grounds dialogue generation in explicitly referenced scenarios and personas derived from teachable moments, whereas DiaSynth generates discussion points and personas, and creates a suitable triggering event based on the discussion point without external guidance. SafeTalkCoach’s explicit scenario conditioning promotes greater contextual variation. DiaSynth, by contrast, imposes minimal constraints on personas and behaviors, resulting in marginally higher strategy and trajectory diversity, but at a cost: reduced constraints weaken control over dialogue focus, making DiaSynth more prone to off-topic generation as task complexity increases. SafeTalkCoach accepts this tradeoff, prioritizing controllability, topical focus, and guideline grounding over maximal scalability, while still achieving strong dialogue-level diversity.

Compared with FCS, SafeTalkCoach achieves higher situational coverage (4.47 vs. 4.20), showing the value of explicitly specifying scenarios to diversify conversational contexts. FCS, however, scores higher on parental strategy and dialogue trajectory diversity. This is because it tends to generate excessively long parent turns (140 vs. 47) (Table \ref{tab:dataset_comparison}), 
% Although these exchanges may appear detailed and step-by-step, they often read as lecture-style monologues rather than natural dialogue. Moreover, the framework often produces highly open and engaged child responses, leading to unrealistic, prolonged question-answering and negotiation. 
which increases parental strategy diversity by enabling detailed and step-by-step guidance, but at the cost of lecture-like monologues that violate best practices. Moreover, the framework often produces highly open and engaged child responses, leading to prolonged question-answering and negotiation that creates diverse dialogue trajectory but does not resemble real parent-child interactions. 
% This highlights the limitation of simulation-based frameworks in modeling \textbf{resistant or suboptimal behaviors without external control mechanisms}. 
SafeTalkCoach, in contrast, sacrifices maximum strategy and trajectory diversity for guideline adherence and realistic engagement modulation, which is a deliberate design choice for practitioner utility. 
% New behaviors can easily be added to the engagement score module to further enhance trajectory diversity if needed.

\begin{table}[t]
\centering
\footnotesize
\hspace*{-0.8cm}
\begin{subtable}[t]{\linewidth}
\centering
\setlength{\tabcolsep}{4pt}
\renewcommand{\arraystretch}{1.1}

\begin{tabular}{lcccccc}
\toprule
& \multicolumn{3}{c}{\textbf{Realism}} 
& \multicolumn{3}{c}{\textbf{Comm. Quality}} \\
\cmidrule(lr){2-4}
\cmidrule(lr){5-7}

\textbf{Framework}
& Nat.
& Coh.
& Contr.$\downarrow$
& Warmth
& Info.
& Engage \\

\midrule
STC
& 2.98
& 3.00
& 6.24\%
& 4.88
& 4.81
& \textbf{4.87} \\

Direct
& 3.00
& 3.00
& \textit{18.31\%}
& 4.10
& 4.16
& 3.98 \\

Procedural
& 3.00
& 3.00
& 9.78\%
& 4.87
& 4.81
& 4.71 \\

PSYDIAL
& 3.00
& 3.00
& 7.29\%
& 4.95
& \textbf{4.89}
& 4.78 \\

DiaSynth
& 3.00
& 3.00
& 10.35\%
& 4.67
& 4.36
& 4.50 \\

FCS
& \textit{2.74}
& 2.99
& \textbf{2.32\%}
& \textbf{4.96}
& 4.81
& 4.75 \\

\bottomrule
\end{tabular}
\end{subtable}

\caption{Evaluation results for realism and communication quality. $\downarrow$ denotes that smaller values are preferred.}
\label{tab:geval_realism_comm_quality}
\end{table}

\paragraph{Realism}
LLM-based naturalness and coherence scores saturate for most frameworks (2.98–3.00), limiting the metric's discriminative power. The exception is FCS, which scores lower on naturalness (2.74) due to its verbose parent utterances, underscoring that multi-agent simulation requires explicit length and style guidance to remain natural. Contradiction rates reveal additional differences in coherence. Most approaches are under or around 10\%, while Direct Prompting is markedly higher (18.31\%), reflecting weaker turn-level consistency when many dialogues are generated at once. Turn-by-turn simulation, i.e., SafeTalkCoach (6.24\%) and FCS (2.32\%), are more consistent, suggesting that incremental simulation introduces structure that facilitates local coherence. SafeTalkCoach further benefits from explicit guidance, combining strong naturalness with low contradictions.

\paragraph{Communication Quality} 
SafeTalkCoach scores highest in Engagement (4.87) and competitively on Warmth (4.88) and Information (4.81). The Engagement result suggests SafeTalkCoach's engagement module successfully conditions parent behavior on child responsiveness, producing adaptive rather than scripted interactions. Most frameworks score relatively high in delivering warmth (4.87–4.96) and helpful information (4.81-4.89), indicating that LLMs internalize supportive parenting norms even without explicit instructions. By contrast, direct prompting performs noticeably worse in all three dimensions, as it tends to produce shallow, emotionally neutral conversations that prioritize facts over emotional exchange. Likewise, DiaSynth’s diverse dialogues score lower in communication quality, reflecting the limits of self-contained, unguided frameworks when domain-specific guidance is required. These results validate SafeTalkCoach's design, which features a structured, guideline-grounded approach that is particularly effective in supporting high-quality, engagement-promoting communications.

\subsection{Human Evaluation of Communication Mistakes}

To complement communication quality evaluation, we conduct a human analysis on a subset of generated dialogues by identifying common practical mistakes as suggested by practitioners \cite{eekahost_common_2019, kaye_five_2022, elledge_5_2017}, which includes giving too much information \textit{(TMI)}, not admitting awkwardness \textit{(NAW)}, assuming teens’ knowledge \textit{(ATK)}, using fear/shame/judgment/absolutes \textit{(FSJ)}, missing a teachable moment \textit{(MTM)} and not answering questions \textit{(NAQ)}. It is worth mentioning that these mistakes are independent of guidelines used in generation to eliminate potential circularity in the assessment. We manually examine the presence of mistakes and report overall and per-mistake presence rates, as detailed in Appendix \ref{appx:human_evaluation}.

SafeTalkCoach exhibits the lowest overall mistake rate across all systems (Table \ref{tab:human_quality}), indicating stronger adherence to best practices. Although all frameworks show competence in acknowledging awkwardness and avoiding fear, shame, judgment, and absolutes (likely reflecting parenting knowledge is well represented in LLM training data), SafeTalkCoach shows marked advantages in avoiding assumptions about teen knowledge and utilizing teachable moments. This finding suggests that explicit grounding in sexual health guidelines and scenarios enables behaviors beyond those internalized during pretraining. Notably, all FCS-generated dialogues exhibit TMI errors, consistent with our earlier finding of excessive parent utterance length, whereas SafeTalkCoach maintains higher concision and avoids information overload.

One limitation emerged in conversation terminations was that a subset of SafeTalkCoach dialogues contains unanswered child questions near dialogue end points, although we instruct the moderator to terminate when the child has no further questions. These results highlight an inherent tension in multi-agent architectures, which rely more heavily on external control structures than internal coherence cues typical of single-model generation. This finding motivates refinement of our stopping criteria. 

\begin{table}[t]
\centering
\footnotesize
\hspace*{-0.8cm}
\begin{subtable}[t]{\linewidth}
\centering
\setlength{\tabcolsep}{4pt}
\renewcommand{\arraystretch}{1.1}

\begin{tabular}{lccccccc}
\toprule
\textbf{Framework}
& TMI & NAW & ATK & FSJ & MTM & NAQ & OVR \\

\midrule

STC
& 0.30
& \textbf{0.00}
& \textbf{0.00}
& 0.10
& \textbf{0.00}
& 0.20
& \textbf{0.10} \\

Direct
& \textbf{0.00}
& 0.15
& 0.65
& 0.20
& 0.95
& \textbf{0.00}
& 0.33 \\

Procedural
& 0.05
& 0.10
& 0.55
& 0.15
& 0.95
& \textbf{0.00}
& 0.30 \\

PSYDIAL
& 0.25
& 0.05
& 0.60
& 0.05
& 0.45
& \textbf{0.00}
& 0.23 \\

DiaSynth
& 0.05
& 0.05
& 0.75
& 0.30
& 0.65
& \textbf{0.00}
& 0.30 \\

FCS
& \textit{1.00}
& \textbf{0.00}
& 0.75
& \textbf{0.00}
& 0.65
& \textbf{0.00}
& 0.40 \\

\bottomrule
\end{tabular}
\end{subtable}

\caption{Evaluation results for communication mistakes}
\label{tab:human_quality}
\end{table}

We additionally conduct ablation studies to evaluate the effectiveness of each component. Full details are provided in Appendix \ref{appx:ablation}.

%% file: sections/5_conclusion.tex
\section{Conclusion}
\label{conclusion}
In this paper, we introduce SafeTalkCoach, a multi-agent dialogue simulation framework for parent-child conversations about sexual health. By integrating crowd-sourced and synthetic scenarios, evidence-based guidelines, and agent-level control modules, SafeTalkCoach produces dialogues that are realistic and aligned with evidence-based best practices. Our hierarchical diversification mechanism, operating across data, architectural, and prompting levels, enables broad situational coverage while avoiding unrealistic behavior, low controllability, and off-topic risks. 

SafeTalkCoach and the accompanying dataset offer practical resources for research on sensitive health communications, with potential applications in simulation-based training and automated intervention design. Future work includes adding more discussion topics, conducting expert and parent evaluations, and developing automated training platforms that offer parents low-stakes practice opportunities. We hope this framework and the dataset support both AI research and health communication practices in real-world settings.

%% file: sections/limitations.tex
\section*{Limitations}
\label{limitations}
This study has several limitations. First, we consider only three topics (safe sex, abstinence, consent), which limits generalizability. Our hybrid data collection pipeline also does not replace human annotation: LLM-generated scenarios, even when guided by teachable moments, may diverge from the situations parents actually encounter, and crowd-sourced scenarios may have different biases, such as self-selection bias. Future work will expand topical coverage and collect scenarios through surveys or interviews with parents to enhance authenticity and representativeness.

Second, our persona and control modules simplify some determinants of real parent–child conversations that are difficult and unreliable to model uniformly with LLMs, such as relationship closeness, emotional stakes, and family communication habits. The multi-agent architecture can also miss natural stopping points, occasionally leaving child questions unanswered. Future work will incorporate richer interaction factors, stronger termination criteria, and deeper personalization. More broadly, agent behavior is sensitive to prompts and model choice, which may impact reproducibility; systematic analyses of prompt and model sensitivity remain for future work.

Finally, evaluation remains challenging. Privacy and ethical constraints limit access to ground-truth parent–child sexual health dialogues, increasing reliance on LLM-as-a-judge methods, which may exhibit biases (e.g., order and self-agreement effects). Although we included a focused human evaluation, we lack extensive feedback from domain experts and parents, as well as downstream validation. We plan larger-scale expert and parent evaluations, followed by user studies to assess ecological validity and pedagogical value in future work.

%% file: sections/ethical_considerations.tex
\section*{Ethical Considerations}
\label{ethical_considerations}

Designed for the sensitive domain of parent-child conversations about sexual health, a few aspects should be considered to ensure ethical use of the SafeTalkCoach framework and the dataset. First, \textbf{SafeTalkCoach is never meant to replace professional counseling or medical services, or provide any medical advice}. As LLMs can provide false, misleading responses or unsafe instructions, we recommend that any use of the SafeTalkCoach framework or dataset be accompanied by factual knowledge about sexuality and authoritative communication guidelines, or be supervised by experts. Second, as noted in the limitations, the data source used in this study may not be representative or comprehensive. As a result, cross-cultural validation and fairness vetting should be conducted before adopting the framework and dataset on a large scale. Third, we extract only publicly available posts from Reddit and StackExchange in our data collection pipeline. As inspected by the two annotators, no Personally Identifiable Information is present in the content of selected posts. To further mitigate privacy issues, the original posts are safely discarded after extracting the high-level descriptions and personas. 

%% file: sections/ablation_setup.tex
\section{Ablation Study}
\label{appx:ablation}
To better understand the contributions of different components, we conduct a series of ablation studies. We sample 40 scenarios from our data source, where 6 scenarios are crowd-sourced, and 34 scenarios are synthetic, based on their proportions. For each ablation, we generate three dialogues per scenario, resulting in a set of 120 dialogues. We then use the same setup as our main experiments to compare dialogues generated by each ablation.

\subsection{Ablation Study on Diversity Mechanism}
We ablate individual diversity mechanisms to examine their contributions to the diversity of dialogues. 

We define the following ablations:

\begin{itemize}
    \item \textbf{Full Functionality}, where all the implemented features are present in the framework.
    \item \textbf{No Scenario}, where we only provide parent and child personas, but no scenarios. 
    \item \textbf{No Planning Module}, where the parent agent does not have the planning module.
    \item \textbf{No engagement Score}, where no engagement modulation is used by the child agent. 
    \item \textbf{No Upper Limit}, where no $\mu_{max}$ is set for the engagement score module.
\end{itemize}

We evaluate the diversity of each ablation using the LLM-based evaluation of dialogue-level diversity as discussed in \ref{experiment_setup}. The results are shown in Table \ref{tab:ablation-winrate}. Full system implementation is consistently preferred in LLM-based diversity evaluation over variants that removes scenarios, the planning module, or the engagement score module, demonstrating the contribution of each component. Although individual components target specific dimensions, ablating any one of them decreases overall diversity, suggesting strong interdependence of these dimensions. In contrast, adding $\mu_{max}$ in the engagement score module does not introduce meaningful improvement of diversity. This is likely because the engagement score sampling process is already sufficient in diversifying dialogue trajectories. 

\subsection{Ablation Study on Key Components}
We ablate key components of the system to evaluate their impacts on the realism and communication quality of the dialogues. 

We define the following ablations: 
\begin{itemize}
    \item \textbf{Full Functionality}, where all the implemented features are present in the framework.
    \item \textbf{Basic Parent Agent}, where the parent agent only takes its persona and basic instructions as input. No planning module or guidelines are involved.
    \item \textbf{Basic Child Agent}, where the child agent only takes its persona and basic instructions as input. No engagement modulation is involved. 
    \item \textbf{No Moderator Agent}, where no moderator is involved, and the dialogue only terminates after a certain number of turns.
\end{itemize}

We evaluate the realism and communication quality of each ablation using LLM-based evaluation as discussed in \ref{experiment_setup}. As shown in Table \ref{tab:ablation_geval}, naturalness drops when control modules for parent, child, or moderator agent are removed, with the largest drop occurring when ablating on moderator. This further validates our earlier finding that multi-agent simulation framework requires carefully crafted external control mechanisms on conversation termination. In addition, removing planning module reduces the ability of parent agent to convey helpful information, and removing engagement score module from the child agent also weakens the parent agent's ability to encourage engagement, reinforcing the importance of prior planning and engagement-aware adaptation.

\begin{table}[t]
\centering
\footnotesize
\setlength{\tabcolsep}{3pt}
\renewcommand{\arraystretch}{1.1}
\begin{tabular}{lcccc}
\toprule
& \multicolumn{2}{c}{No Scenario}
& \multicolumn{2}{c}{No Planning Module} \\
\cmidrule(lr){2-3}
\cmidrule(lr){4-5}

Dimension
& Full & Variant
& Full & Variant \\
\midrule

Overall
& 83.3\% & 8.3\%
& 50.0\% & 16.7\% \\

Situational
& 91.7\% & 0.0\%
& 41.7\% & 8.3\% \\

Strategy
& 83.3\% & 8.3\%
& 50.0\% & 16.7\% \\

Trajectory
& 83.3\% & 8.3\%
& 41.7\% & 16.7\% \\

\addlinespace[4pt]
\midrule
\addlinespace[2pt]

& \multicolumn{2}{c}{No Engagement Score}
& \multicolumn{2}{c}{No Upper Limit} \\
\cmidrule(lr){2-3}
\cmidrule(lr){4-5}

Dimension
& Full & Variant
& Full & Variant \\
\midrule

Overall
& 75.0\% & 25.0\%
& 33.3\% & 25.0\% \\

Situational
& 83.3\% & 16.7\%
& 41.7\% & 33.3\% \\

Strategy
& 75.0\% & 25.0\%
& 33.3\% & 25.0\% \\

Trajectory
& 75.0\% & 16.7\%
& 0.0\%  & 33.3\% \\

\bottomrule
\end{tabular}
\caption{Ablation study results on diversity mechanisms reporting win rates of \textbf{Full Functionality (\textit{Full})} against each \textbf{ablation variant (\textit{Variant}).}}
\label{tab:ablation-winrate}
\end{table}

\begin{table}[t]
\centering
\footnotesize
\begin{subtable}{\columnwidth}
\centering
\setlength{\tabcolsep}{4pt}
\renewcommand{\arraystretch}{1.0}

\begin{tabular}{lccccc}
\toprule
& \multicolumn{2}{c}{\textbf{Realism}} 
& \multicolumn{3}{c}{\textbf{Communication Quality}} \\
\cmidrule(lr){2-3}
\cmidrule(lr){4-6}

\textbf{Variant}
& Nat.
& Coh.
& Warmth
& Info.
& Engage \\

\midrule
Full Func.
& 2.97 & 3.00
& 4.80 & 4.71 & 4.78 \\

Basic Parent
& \textit{2.89} & 3.00
& 4.90 & 4.65 & 4.89 \\

Basic Child
& \textit{2.90} & 3.00
& 4.88 & 4.62 & 4.73 \\

No Moderator
& \textit{2.82} & 3.00
& 4.98 & 4.86 & 4.95 \\

\bottomrule
\end{tabular}
\end{subtable}

\caption{Ablation study results on key components. Realism dimensions are rated on a 1-3 scale, while communication quality dimensions are rated on a 1-5 scale.
All values are reported as mean scores.}
\label{tab:ablation_geval}

\end{table}

%% file: sections/engagement_score.tex
\section{Engagement Score Computation Details}
\label{appx:engagement}

The Engagement Score Module maintains a 4-level engagement score, where $\{0, 1\}$ represents disengagement, and $\{2, 3\}$ represents engagement. At each turn $t$, the engagement score $e_t$ is sampled from a Gaussian distribution as follows:
\begin{equation}
e_t \sim \mathcal{N}(\mu_t, \sigma^2)
\end{equation}

Where $\mu_t$ denotes the mean of the engagement distribution at turn $t$, i.e., the child’s most probable internal engagement state. $\sigma^2$ is the variance of the Gaussian distribution, controlling the stochasticity of engagement across turns. We initialize $\mu_t = 0$ for child agents with \textit{dismissive or resistant} attitudes, and $\mu_t = 1$ for child agents with \textit{somewhat comfortable} attitudes, before the start of the dialogues. We also set $\sigma= 0$ for the first turn of the dialogues, and $\sigma = 0.5$ in subsequent turns, to maintain consistency with personas.

 Before sampling at each turn $t$, the agent rates the parent’s utterance along dimensions that may influence engagement. Each dimension is assigned a weight determined through experiments: positive factors include \textit{presence of open-ended questions ($w = 0.5$), caring ($w = 0.25$)}, and \textit{empathy ($w = 0.25$)}, and negative factors include the \textit{presence of hostile language ($w = -0.33$), embarrassment ($w = -0.33$)}, \textit{and poor communication skills ($w = -0.33$)}. The weighted ratings are then transformed to a scale that matches the scale of engagement scores. This yields an estimate of the most probable engagement score after hearing the parent’s utterance under ideal conditions, $\hat{e}_t$. The new mean of the engagement score is then computed using a smooth linear rule:
\begin{equation}
\mu_t = (1 - \alpha)\,\mu_{t-1} + \alpha\,\hat{e}_t
\end{equation}

Where $\mu_t$ is the updated mean engagement score at turn $t$, $\mu_{t-1}$ is the engagement mean from the previous turn, $\hat{e}_t$ is the estimated ideal engagement score inferred from the perceived ratings of the parent’s utterance after normalization to the engagement scale, and $\alpha \in (0,1]$ is a smoothing coefficient controlling the influence of the current utterance on engagement. We set $\alpha$ to 0.3 in our framework based on experiments.

We also introduce $\mu_{\max}$, a random maximum possible mean engagement score for each child persona. We set $\mu_{\max} \sim \mathcal{U}\!\left(1.0,\; 1.8\right)$ for child agents with \textit{dismissive or resistant} attitudes, and $\mu_{\max} \sim \mathcal{U}\!\left(2.0,\; 2.8\right)$ for child agents with \textit{somewhat comfortable} attitudes, based on experiments.

To prevent unrealistic behaviors caused by sampling, the engagement score can change by at most 2 for each turn.

%% file: sections/data_collection.tex
\section{Crowd-sourced Data Collection Details}
\label{appx:collection}
We collected 2,188 posts from subreddits \textit{r/Mommit, r/Daddit, r/Parenting, r/AskParents,} and \textit{r/Teenagers} on Reddit, and 400 posts from the site \textit{parenting.stackexchange.com} on Stack Exchange using their official APIs, due to high relevance to the discussion topics. 
We searched for the following relevant keywords for each topic: 
\begin{itemize}
    \item Safe Sex: \textit{"safe sex", "protection", "condom", "contraception",
"birth control", "STI", "pregnancy scare"}.
    \item Abstinence: \textit{"abstinence", "no sex", "not ready", "waiting", "celibacy"}.
    \item Consent: \textit{"consent", "boundaries", "pressure", "harassment", "assault"}.
\end{itemize}
To develop an instruction for filtering out relevant and high-quality posts in the absence of clear definitions or rubrics, two authors who are graduate students with experience in qualitative analysis inspected and annotated 58 random posts separately on whether a post is suitable for a conversation and which topics it is relevant to, in order to explore possible criteria that should be taken into consideration. After initial annotation, they discussed potential criteria of suitableness and relevance to topics, and updated their own annotation based on the discussion, yielding a Cohen's \(\kappa = 0.88\). They then created an instruction (Table \ref{tab:instructions}) that both agree upon, which was provided to GPT-5 mini to perform filtering on the same set of 58 posts, resulting in a recall of 75\%, which demonstrates the effectiveness of the instruction in detecting relevant posts with low false negative rates.

The LLM then performed an initial screening based on the instructions and filtered out 293 posts. The annotators then divided the posts into two halves (one with 146 posts and the other with 147 posts), and manually inspected them following the same instructions, resulting in 74 high-quality posts. Since many posts clearly reveal certain aspects of the parent and child personas, such as gender and age, and the teachable moment with minimal ambiguity and subjectivity, the annotators also labeled the parent and child personas, as well as the corresponding category of the teachable moments, if explicitly revealed in the posts. Notes were left on certain long posts to help LLM extract the scenario. 

The labeled topics, personas, teachable moments, and notes were then fed into GPT-4o mini (with temperature 0.8 to maintain consistency with scenario synthesis) to generate a scenario description for each post. Information that cannot be extracted from the post was assigned randomly. All scenario descriptions and personas generated are anonymous, ensuring that our dataset does not contain sensitive information that could be misused. 

%% file: sections/baseline_setup.tex
\section{Baseline Setup Details}
\label{appx:baseline}
We discuss our setup details for each baseline and the statistics of the data used in our experiments. For our implemented baselines (Direct Prompting and Procedural Prompting), we used GPT-4o mini (temperature 0.7). For established baselines (PSYDIAL, DiaSynth, FCS), we used GPT-4o mini and retained their original temperatures. 
\begin{itemize}
    \item \textbf{Direct Prompting}. For each combination of parent persona, child persona, and the discussion topic, we instruct the LLM to generate 15 dialogues in a single prompt. 
    \item \textbf{Procedural Prompting}. For each combination of parent persona, child persona, and discussion topic, we instruct the LLM to generate 15 conversation starters for each pair and use separate prompts to generate one dialogue per conversation starter. 
    \item \textbf{PSYDIAL}. In this framework, the extroversion dimension of the two persons' personalities are first randomly sampled, then fed into the LLM to generate a dialogue. We modify this framework to utilize our defined parent and child personas, rather than general personality types. For each discussion topic, we generate 350 dialogues.
    \item \textbf{DiaSynth}. The framework first generates subtopics and corresponding personas for each topic, then generates dialogues between each pair of personas by Chain-of-Thought (CoT) reasoning. We restrict the framework to only generate parent and child personas, and only pair parent with child during dialogue generation. During our experiments, we noticed that when the number of subtopics is large, the framework fails to stay within the context of sexual health and generates an excessive number of dialogues that are completely off-topic, which affects the validity of our evaluation approaches. As a result, we use a small number of 6 subtopics per topic, and 64 pairs of parent and child personas for each subtopic to ensure a reasonable number of dialogues for comparison. 
    \item \textbf{Family Conversation Simulation}. The framework uses direct role-play between parent and child agents. We adopt our parent personas rather than the general parenting styles, and add child personas and discussion topics. For each combination of parent persona, child persona, and discussion topic, we generate 15 dialogues using the zero-shot full-context version of the framework.
\end{itemize}

During our experiment, we identified that some frameworks, despite being instructed to generate a certain number of dialogues, subtopics, or personas, generate a wrong number of responses. As a result, we keep the minimum number of dialogues generated by any frameworks, and down-sample other results to comparable number to ensure a fair comparison. Detailed statistics of data used in our experiments are shown in Table \ref{tab:dataset_comparison}.

%% file: sections/human_annotation_process.tex
\section{Human Evaluation Annotation Process}
\label{appx:human_evaluation}

Due to the labor-intensive nature of detailed mistake annotation, we randomly sampled 20 dialogues from each framework (120 total). To mitigate biases, we removed the source of each dialogue and randomly permuted them. Two authors, graduate students with experience in qualitative analysis, independently labeled a random 10\% subset (12 dialogues). The initial inter-annotator agreement yielded Cohen's \(\kappa = 0.28\), reflecting the inherent subjectivity of the task. Following discussion to resolve disagreements on subjective aspects of communication mistakes, the annotators collaboratively developed operational definitions and guidelines for each mistake category (Table \ref{tab:instructions_mistakes}). A second 10\% subset was then independently annotated to assess the effectiveness of the guidelines, achieving a Cohen's \(\kappa = 0.70\), indicating substantial agreement \cite{landis_measurement_1977}. The remaining 80\% of dialogues were evenly divided between the two annotators for independent annotation. The labels of two annotators are then combined to calculate the presence rate of each mistake and the overall presence rate across frameworks.

%% file: sections/dataset_statistics.tex
\begin{table}[t]
\centering
\small

\begin{tabular}{l r}
\toprule
\textbf{Category} & \textbf{Count / Value} \\
\midrule
\multicolumn{2}{l}{\textbf{Overall Statistics}} \\
\midrule
Total number of dialogues & 1495 \\
Total number of utterances & 10520 \\
\quad Parent utterances & 5260 \\
\quad Child utterances & 5260 \\
\midrule
\multicolumn{2}{l}{\textbf{Average Lengths}} \\
\midrule
Average parent utterance length (words) & 47.37 \\
Average child utterance length (words) & 21.09 \\
Average dialogue length (utterances) & 7.04 \\
\midrule
\multicolumn{2}{l}{\textbf{Dialogues by Parent Role}} \\
\midrule
Father & 733 \\
Mother & 762 \\
\midrule
\multicolumn{2}{l}{\textbf{Dialogues by Parent Attitude}} \\
\midrule
Progressive & 753 \\
Traditional & 742 \\
\midrule
\multicolumn{2}{l}{\textbf{Dialogues by Child Attitude}} \\
\midrule
Dismissive or Resistant & 715 \\
Somewhat Comfortable & 780 \\
\midrule
\multicolumn{2}{l}{\textbf{Dialogues by Child Age}} \\
\midrule
11 & 168 \\
12 & 144 \\
13 & 171 \\
14 & 153 \\
15 & 171 \\
16 & 211 \\
17 & 243 \\
18 & 234 \\
\midrule
\multicolumn{2}{l}{\textbf{Dialogues by Child Gender}} \\
\midrule
Female & 750 \\
Male & 745 \\
\midrule
\multicolumn{2}{l}{\textbf{Dialogues by Topic}} \\
\midrule
Abstinence & 462 \\
Consent & 475 \\
Safe sex & 558 \\
\midrule
\multicolumn{2}{l}{\textbf{Dialogues by Teachable Moments}} \\
\midrule
\multicolumn{2}{l}{\textit{Safe Sex}} \\
\quad Major Life Events & 201 \\
\quad Social Media & 117 \\
\quad Mass Media & 111 \\
\quad Everyday Occurrences & 129 \\
\midrule
\multicolumn{2}{l}{\textit{Consent}} \\
\quad Major Life Events & 120 \\
\quad Social Media & 105 \\
\quad Mass Media & 102 \\
\quad Everyday Occurrences & 148 \\
\midrule
\multicolumn{2}{l}{\textit{Abstinence}} \\
\quad Major Life Events & 144 \\
\quad Social Media & 96 \\
\quad Mass Media & 102 \\
\quad Everyday Occurrences & 118 \\
\bottomrule
\end{tabular}
\caption{SafeTalkCoach Dataset statistics.}
\label{tab:dataset_statistics}
\end{table}

\begin{table*}[t]
\centering
\small
\begin{tabular}{l r r r r r r}
\toprule
\textbf{Metric} & \textbf{Direct} & \textbf{Proc.} & \textbf{PSYDIAL} & \textbf{DiaSynth} & \textbf{FCS} & \textbf{STC}  \\
\midrule
Dialogues & 750 & 750 & 750 & 747 & 750 & 750 \\
Total utterances & 3760 & 6882 & 9215 & 9469 & 9000 & 5188 \\
\quad Parent & 2142 & 3708 & 4529 & 4859 & 4500 & 2594 \\
\quad Child & 1618 & 3174 & 4686 & 4610 & 4500 & 2594 \\
\midrule
Avg parent utterance length (words) & 11.43 & 24.83& 30.93& 18.76& 140.15& 47.29 \\
Avg child utterance length (words) & 7.02 & 13.48& 15.95& 12.77& 69.05& 21.16 \\
Avg dialogue length (utternaces) & 5.01 & 9.18& 12.29& 12.68& 12.0 & 6.92 \\
\bottomrule
\end{tabular}
\caption{Dataset statistics for evaluation across frameworks.}
\label{tab:dataset_comparison}
\end{table*}

%% file: sections/dialogue_examples.tex
\begin{table*}[p]
\centering
\scriptsize
\renewcommand{\arraystretch}{0.95}

\begin{tabular}{|m{0.18\linewidth}|m{0.78\linewidth}|}
\hline

\textbf{Example 1} &
\ 

\textbf{Parent}	Hey sweetie, can we sit down for a minute? I heard about that incident at school where a boy touched someone inappropriately, and I want to talk to you about consent and what it means for your safety and feelings. It’s really important to me that you know your body is yours, and you have every right to set boundaries with others. What do you think about that?

\textbf{Child}	Um, I guess I get it. Like, it’s important to say no if someone makes you uncomfortable, right?

\textbf{Parent}	Yes, exactly! Saying no is really important when something doesn’t feel right. It’s okay to speak up if you’re uncomfortable, and you can always come to me or another trusted adult if you need help. What else do you think about this?

\textbf{Child}	Yeah, I get that. It’s just hard sometimes to know when to say no, especially if the person is a friend or someone you know.

\textbf{Parent}	I know it can be tricky when it's someone you know, but remember that your feelings are what matter most. If something feels off or makes you uncomfortable, it's totally okay to say no, even to a friend. Have you talked to anyone else about how you feel regarding this?

\textbf{Child}	I haven't really talked to anyone else about it. I guess I just thought it was super weird and kinda scary.
\vspace{0.2cm}
\\
\hline

\textbf{Example 2} &
\ 

\textbf{Parent}	Hey, I overheard you talking about someone sending explicit messages without asking first. That kind of situation can really mess with someone's feelings, and it made me think we should chat about consent. What do you know about consent and why it's important?

\textbf{Child}	Can we not talk about this right now? It's kinda awkward.

\textbf{Parent}	I get that it feels awkward, but it's really important. Consent means making sure both people are okay with something happening, whether it's sending messages or anything else. What do you think about how someone might feel if they get messages like that without wanting them?

\textbf{Child}	I guess it could make someone feel really uncomfortable or even scared if they get messages like that. It's just... kinda hard to talk about, you know?

\textbf{Parent}	I totally understand it feels awkward, but talking about it can really help. When someone gets a message like that without asking, it can really shake them up, and that's why respecting boundaries is so important. Have you thought about how you would want to handle it if someone sent you something like that?

\textbf{Child}	I mean, I guess I'd feel pretty weird about it. I wouldn't really know how to respond. It's just awkward to think about.

\textbf{Parent}	I totally get that it feels awkward, but it’s important to talk about how to handle things like that. If you ever get a message that makes you uncomfortable, you have every right to ignore it or let someone know. It's about protecting your feelings and respecting your own boundaries. How do you think you'd want to respond if it happened to you?"

\textbf{Child}	I guess I’d just try to ignore it or maybe talk to a friend about it. It just feels weird to think about.

\\
\hline

\textbf{Example 3} &
\ 

\textbf{Parent}	Hey sweetie, I overheard your friends talking about feeling pressured to share secrets, and it got me thinking about how important it is to talk about consent, especially when it comes to sharing personal stuff. What do you think about that?

\textbf{Child}	Yeah, I guess it’s important. But it’s kind of awkward to talk about.

\textbf{Parent}	I get that it can feel awkward, but it’s really important to talk about. Consent isn’t just about saying yes or no to things like dating; it also means you get to choose what personal stuff you share and with whom. What do you already know about consent in that way?

\textbf{Child}	I don’t know, Mom. Can we just talk about something else?

\textbf{Parent}	I understand it feels awkward, and that's okay. Consent is really about feeling comfortable with what you share and knowing it's perfectly fine to keep some things private. Have you ever felt pressured to share something you didn't want to?

\textbf{Child}	I guess I've felt that way a bit, like when friends expect me to share stuff just because they are. It can be kind of weird.

\textbf{Parent}	I totally understand that it can feel weird, but it’s really important to recognize that you don’t have to share anything you’re not comfortable with. It’s okay to say no or to keep things private, and true friends will respect that. How do you think you could handle a situation where you feel pressured to share something?

\textbf{Child}	I guess I could just say I’m not comfortable sharing that, or maybe change the subject lightly. It’s tough sometimes, though.

\\
\hline

\end{tabular}

\caption{SafeTalkCoach Dialogue Examples. The discussion topic is \textit{consent}.}
\label{tab:safetalkcoach_examples}
\end{table*}

\begin{table*}[p]
\centering
\scriptsize
\renewcommand{\arraystretch}{0.95}

\begin{tabular}{|m{0.18\linewidth}|m{0.78\linewidth}|}
\hline

\textbf{Example 1} &
\ 

\textbf{Parent}	What do you think about how society talks about consent?

\textbf{Child}		I think it can be confusing. Sometimes it seems like people don't take it seriously.

\textbf{Parent}	That’s true. We need to change that by having open conversations, like we’re doing now.

\textbf{Child}		Yeah, it helps to talk about it openly.

\\
\hline

\textbf{Example 2} &
\ 

\textbf{Parent}	Have you heard about boundaries in relationships?

\textbf{Child}		Kind of, but not really.

\textbf{Parent}	Boundaries are limits we set for ourselves about what we are comfortable with. It’s really important to communicate those with others.

\textbf{Child}		So, it’s okay to say no if I don’t want to?

\textbf{Parent}	Absolutely! Saying no is a healthy part of boundaries.

\\
\hline

\textbf{Example 3} &
\ 

\textbf{Parent}	I want to talk about what to do if someone tries to pressure you.

\textbf{Child}		Like what kind of pressure?

\textbf{Parent}	Any kind—like trying to make you do something you're not ready for.

\textbf{Child}		I’d probably feel really uncomfortable.

\textbf{Parent}	And that's okay! You can always say no and walk away.

\\
\hline

\end{tabular}

\caption{Direct Prompting Dialogue Examples. The discussion topic is \textit{consent}.}

\label{tab:direct_examples}
\end{table*}

\begin{table*}[p]
\centering
\scriptsize
\renewcommand{\arraystretch}{0.95}

\begin{tabular}{|m{0.18\linewidth}|m{0.78\linewidth}|}
\hline

\textbf{Example 1} &
\ 

\textbf{Child}	Hey Mom, can we talk about something? I’ve been thinking a lot about consent lately, especially with everything that’s been going on in school and online. It seems really important, you know?

\textbf{Parent}	Of course, sweetie! I'm really glad you brought this up. Consent is super important, and it’s great that you’re thinking about it. It’s all about making sure that both people in any kind of relationship or interaction feel comfortable and respected. 

Do you have specific questions or situations in mind that you want to talk about? I’m here to listen and help you sort through your thoughts

\textbf{Child}	Yeah, I guess I’ve just been wondering about how to actually communicate consent, you know? Like, it’s one thing to understand it, but when you’re in a situation, it can feel a bit awkward. How do you make sure both people are on the same page without it being super weird?

\textbf{Parent}	I totally get that, and it’s a really important question! Communicating consent doesn’t have to be awkward, and it can actually help both people feel more comfortable. A good way to approach it is to keep the conversation open and relaxed. You can start by checking in with each other, asking questions like, “How are you feeling about this?” or “Is this okay with you?” 

It’s all about being honest and attentive to each other’s feelings. You can also use body language as a guide—if someone seems hesitant or unsure, it’s important to pause and talk about it. Remember, consent is about both people feeling good about what’s happening, and it’s totally okay to stop or change your mind at any time. 

Have you thought about how you might bring this up in a situation?

\textbf{Child}	Yeah, I’ve thought about it a bit. I guess I’d just try to be straightforward and check in like you said. Maybe I could start with something casual, like asking how the other person feels about what’s happening. I think it might help to keep it light, so it doesn’t feel too heavy or serious, you know? 

But I also worry about how the other person might react. What if they feel awkward or don’t know how to respond? I want to make sure it doesn’t turn into a weird moment. How do you think I should handle that?

\textbf{Parent}	I completely understand your concerns, and it’s really thoughtful of you to consider how the other person might feel. It’s true that talking about consent can sometimes feel a bit awkward, but the key is to create a safe space for both of you. 

If you sense that the other person is feeling awkward or unsure, you can reassure them that it’s totally okay to take a moment to think about it. You might say something like, “Hey, if this feels a bit weird, we can just take a break and talk about something else for a bit.” That way, you’re showing that you respect their feelings and are open to whatever they need.

You can also share your own feelings, like saying, “I just want to make sure we’re both comfortable with this.” This way, it feels more like a team effort rather than putting pressure on one person. Remember, it’s all about mutual respect and making sure both of you feel good about what’s happening. 

\textbf{Child}	Yeah, that makes a lot of sense. I like the idea of keeping it a team effort. It feels way better to approach it like that instead of making it all about one person. I guess it’s just about being open and honest, right? 

And I think it’s cool to remind each other that it’s okay to take a step back if things get weird. It’s like creating a chill vibe where both people can feel safe to express what they want or don’t want. 

Do you think there are any other tips or things I should keep in mind when it comes to consent, especially if I’m in a situation where it might get a bit more serious?

\textbf{Parent}	I'm really glad to hear you feel that way! You’re absolutely right—being open and honest is key, and creating a chill vibe makes it easier for both people to share their feelings. 

When it comes to more serious situations, here are a few additional tips to keep in mind about consent:

1. **Check In Regularly**: Just because you’ve talked about consent once doesn’t mean you shouldn’t check in again as things progress. It’s perfectly fine to ask how the other person is feeling as you move forward.

2. **Be Aware of Non-Verbal Cues**: Sometimes, people might not say exactly how they feel, but their body language can tell you a lot. If you notice someone seems tense or unsure, it’s important to pause and ask how they’re feeling.

3. **Know That Consent Can Change**: Just because someone agreed to something earlier doesn’t mean they have to stick with it if they’re feeling uncomfortable.

\textbf{Child}	Yeah, those are really good points. I like the idea of checking in regularly—it makes sense that feelings can change as things go on. It’s kind of like having a little ongoing conversation, right? 

And being aware of non-verbal cues is super important too. Sometimes people might not feel comfortable saying something, but their body language can say a lot. I guess just being attentive to that can help a lot in making sure everyone feels good about what’s happening.

I also think it’s really important to remember that consent isn’t just a one-time thing. Like you said, it can change, and it’s totally okay for someone to say they’re not feeling it anymore. I want to make sure that whoever I’m with knows they can speak up at any time without feeling pressured. 

Do you think there are any other situations where consent can get tricky, or is it mostly just about being aware and communicating?

\textbf{Parent}	I'm really impressed with how much thought you've put into this! You're absolutely right—consent is an ongoing conversation, and being attentive to both verbal and non-verbal cues is key to making sure everyone feels comfortable.

As for situations where consent can get tricky, there are definitely a few things to keep in mind:

1. **Peer Pressure**: Sometimes, friends or peers might unintentionally pressure someone into doing something they're not comfortable with. It’s important to recognize that everyone should feel free to say no without feeling like they have to justify their choice.

2. **Substance Use**: If someone has been drinking or using drugs, their ability to give clear consent can be impaired. It’s crucial to be mindful of this and ensure that both people are in a clear state of mind when discussing consent.

3. **Power Dynamics**: In situations where there’s a power imbalance—like between a teacher and a student, or an older person and a younger person—consent can become complicated

\textbf{Child}	Yeah, those are really important points. I hadn’t thought about how peer pressure could play a role. It’s so true that sometimes friends might not mean to, but they can make someone feel like they have to go along with something, even if they’re not comfortable. I guess it’s really important to have friends who respect each other’s boundaries.

And the part about substance use is super crucial, too. It’s hard to make clear decisions when you’re not fully aware of what’s going on. I think it’s good to have those conversations beforehand, like making sure everyone knows what they’re okay with before any kind of party or gathering.

The power dynamics thing is definitely something to keep in mind as well. It can be really tricky when there’s an age difference or someone in a position of authority. It’s just another reason why open communication is so important, so everyone feels safe to express themselves. 

Do you think there are other ways to help friends feel comfortable saying no

\textbf{Parent}	I'm really glad to hear that you’re thinking about how to support your friends in feeling comfortable saying no. It’s so important to create an environment where everyone feels safe to express their boundaries. Here are a few ideas that might help:

1. **Normalize Saying No**: Make it clear among your friends that saying no is totally okay. You can even joke about it sometimes, like saying, “If you’re not feeling it, just say no! No hard feelings!” This can help lighten the mood and make it easier for everyone to speak up.

2. **Lead by Example**: If you’re open about your own boundaries and comfortable saying no when you need to, it encourages others to do the same. Showing that it’s okay to prioritize your own comfort can really empower your friends.

3. **Check In Regularly**: Just like we talked about with consent, checking in with your friends during social situations can create a supportive atmosphere. 

\\
\hline

\end{tabular}

\caption{Family Conversation Simulation Dialogue Examples. The discussion topic is \textit{consent}.}
\label{tab:fcs_examples}
\end{table*}

\begin{table*}[p]
\centering
\scriptsize
\renewcommand{\arraystretch}{0.95}
\ContinuedFloat

\begin{tabular}{|m{0.18\linewidth}|m{0.78\linewidth}|}
\hline

\textbf{Example 2} &
\ 

\textbf{Child}	Hey Mom, can we talk about something a bit serious for a minute? I’ve been thinking about consent and how important it is, especially with everything we hear about it these days. I feel like it’s a topic that doesn’t get talked about enough. What do you think?

\textbf{Parent}	Of course, sweetie! I'm really glad you brought this up. Consent is such an important topic, and it's great that you're thinking about it. It’s all about understanding that both people involved need to agree to any kind of physical interaction, and that agreement should be clear and enthusiastic. 

It's also crucial to remember that consent can be withdrawn at any time. Just because someone agreed to something earlier doesn’t mean they have to stick with it if they’re feeling uncomfortable. That’s where boundaries come in. It’s all about respecting each other’s feelings and being open to talking about what you both want or don’t want. 

How do you feel about what you’ve been hearing? Are there specific situations or questions you have in mind?

\textbf{Child}	Yeah, I get that. It makes sense that both people need to be on the same page. I guess what I've been hearing a lot is how important it is to really check in with each other, especially if things start to get more serious. 

I mean, I’ve seen some stuff online where people talk about how sometimes one person might feel pressured to go along with something, even if they’re not totally comfortable. It just makes me think about how important it is to really listen to each other and make sure everyone feels safe and respected. 

Do you think it’s hard for people my age to talk about this kind of stuff openly? Like, how do you think we can make it easier for each other?

\textbf{Parent}	You’re absolutely right, and I appreciate how thoughtful you’re being about this. It can definitely be tough for people your age to talk about consent and boundaries openly. There’s often a lot of pressure, whether it’s from friends, social media, or even just the fear of being judged. 

One way to make it easier is by creating an environment where everyone feels comfortable sharing their thoughts and feelings. It helps to have conversations like this one, where you can express concerns without fear of judgment. Encouraging your friends to talk openly about their feelings and experiences can really make a difference. 

Also, checking in with each other regularly, even in casual situations, can normalize those conversations. It’s about making it a part of your everyday interactions. If you’re ever in a situation where things are getting serious, just asking, “How are you feeling about this?” can go a long way. It shows that you care about their comfort and respect their boundaries.

How do you think you might approach

\textbf{Child}	I think I could start by just being more open with my friends about how I feel. Like, if we're hanging out and the topic comes up, I could bring it up casually and see what they think. Maybe even share my thoughts on how important it is to check in with each other. 

I could also try to lead by example, you know? If I'm in a situation where things might get serious, I could make it a point to ask how they're feeling or if they're comfortable. It might feel a bit awkward at first, but I think it could help make those conversations more normal over time. 

Plus, if I can show my friends that it's okay to talk about this stuff without feeling judged, maybe they'll feel more comfortable doing the same. What do you think?

\textbf{Parent}	I think that’s a fantastic approach! Being open with your friends about how you feel can really help create a culture of trust and understanding. It’s so important to normalize those conversations, and leading by example is a great way to do that. 

When you ask your friends how they’re feeling or if they’re comfortable, it not only shows that you care, but it also encourages them to be honest about their own boundaries and feelings. It might feel a bit awkward at first, but that’s completely normal. The more you practice, the easier it will become.

And remember, consent is all about communication and respect. If everyone feels safe to express their thoughts, it creates a better environment for everyone involved. Plus, it can help prevent misunderstandings and ensure that everyone is on the same page. 

I’m really proud of you for thinking about how to foster these conversations. Do you have any specific situations in mind where you think you might want to try this out?

\textbf{Child}	Thanks, Mom! I appreciate that. I was thinking about when I’m hanging out with friends, especially if we’re watching a movie or something, and things start to get a little more cozy. That could be a good time to check in and see how everyone’s feeling. 

Also, if I ever see someone in a situation where they might seem uncomfortable, I think I could step in and ask if they’re okay or if they want to talk about it. It might help them feel like they have someone looking out for them. 

What do you think about those situations? Do you think they’re good opportunities to bring it up?

\textbf{Parent}	I think those are excellent opportunities to bring up consent and check in with your friends! When you're in a cozy setting, it can be easy for things to progress without everyone fully discussing how they feel. Taking a moment to ask, “Hey, how are you feeling about this?” shows that you care about their comfort and boundaries, which is so important.

And stepping in when you notice someone might be uncomfortable is really commendable. Sometimes people might not feel comfortable speaking up for themselves, so having a friend like you who is willing to check in can make a big difference. It shows empathy and support, and it can help create a safe space for everyone involved.

Just remember that it’s all about being respectful and open. If you approach these conversations with kindness and understanding, it can help everyone feel more at ease discussing their feelings. Do you feel ready to take those steps when the time comes?

\textbf{Child}	Yeah, I think I do feel ready. I mean, it might be a little awkward at first, but I get that it’s really important to make sure everyone feels comfortable. I like the idea of just checking in casually, like it’s no big deal, so it doesn’t feel too heavy or serious. 

I also want to make sure my friends know that it’s okay to say no or change their mind about something. I think if I can show them that I’m open to talking about it, it might help them feel more comfortable too. 

Do you think there are any other ways I can approach it or things I should keep in mind?

\textbf{Parent}	I'm really glad to hear that you feel ready! It sounds like you have a great mindset about approaching these conversations. Keeping it casual is definitely a good way to make it feel less intimidating. 

One thing to keep in mind is the importance of body language and non-verbal cues. Sometimes, people might not verbally express their discomfort, but their body language can tell a different story. Being attentive to how your friends are reacting can help you gauge whether they’re truly comfortable or not.

Also, remember that it’s okay to share your own feelings and experiences too. When you open up about your thoughts on consent and boundaries, it can encourage your friends to do the same. It’s all about creating a two-way dialogue.

Lastly, don’t hesitate to check in with yourself as well. Make sure you’re comfortable with the situation and the people involved. If something doesn’t feel right to you, it’s perfectly okay to speak up or step away.

\textbf{Child}	Yeah, that makes a lot of sense. I think being aware of body language is super important. Sometimes people might seem fine, but you can tell they’re not really into it just by how they’re acting. I’ll definitely keep an eye out for that.

And I like the idea of sharing my own feelings too. It could help my friends feel like they can open up more. It’s kind of like creating a safe space where everyone can be honest about what they want or don’t want.

I guess I just want to make sure that everyone feels respected and comfortable, including myself. It’s good to remember that it’s okay to step back if something feels off. I think that balance of checking in with others and also being true to my own feelings is really key. Thanks for helping me think through all this!

\textbf{Parent}	I'm really proud of you for thinking so deeply about all of this. It’s clear you have a strong sense of what it means to respect both yourself and others. Remember, consent is an ongoing conversation, not just a one-time agreement. It’s about creating an environment where everyone feels safe to express their feelings and boundaries. 

As you navigate these situations, trust your instincts and always prioritize open communication. And if you ever feel uncertain or need to talk about anything, I'm here for you. You're not alone in this, and I'm always ready to listen. Keep being the thoughtful and caring person you are!

\\
\hline

\end{tabular}

\caption{Family Conversation Simulation Dialogue Examples (continued).}
\label{tab:fcs_examples}
\end{table*}

\begin{table*}[p]
\centering
\scriptsize
\renewcommand{\arraystretch}{0.95}
\ContinuedFloat

\begin{tabular}{|m{0.18\linewidth}|m{0.78\linewidth}|}
\hline

\textbf{Example 3} &
\ 

\textbf{Child}	Hey Dad, can we talk about something? I’ve been thinking a lot about consent lately, especially with everything I see online and at school. It feels like it’s super important to understand, you know?

\textbf{Parent}	Of course, I’m glad you brought this up. Consent is incredibly important, and it’s great that you’re thinking about it. It’s not just about saying “yes” or “no,” but really understanding that both people involved need to feel comfortable and respected in any situation. 

You should always feel free to express your boundaries, and it’s equally important to respect the boundaries of others. Remember, consent can be withdrawn at any time, and no one should ever feel pressured into anything they’re not ready for. 

It’s about communication and making sure that both parties are on the same page. If you ever feel unsure or pressured, it’s okay to step back and reassess the situation. Your feelings and comfort matter a lot, and I want you to always prioritize that. What specific things have you been thinking about?

\textbf{Child}	Yeah, I get that. I’ve been thinking about how sometimes people don’t really talk about it, like they just assume everything is okay without actually checking in with each other. Like, at parties or whatever, I see people getting caught up in the moment and not really considering if both people are actually on the same page. 

And then there’s this whole thing with social media, where people might think it’s okay to send certain messages or photos without really thinking about consent. It just seems like there’s a lot of pressure to go along with things sometimes, and I want to make sure I’m being respectful and understanding of others’ boundaries. How do you think I can be more aware of that in real situations?

\textbf{Parent}	I'm really proud of you for thinking critically about these situations. It’s true that many people get swept up in the moment, and that can lead to misunderstandings about consent. One of the best ways to be more aware is to practice open communication. If you're at a party or in any situation where things might get a little heated, take a moment to check in with the other person. It can be as simple as asking, “Are you comfortable with this?” or “How do you feel about what’s happening?”

Also, remember that consent is an ongoing conversation. Just because someone agrees to something at one moment doesn’t mean they’re obligated to continue if they start feeling uncomfortable. You can set a good example by being clear about your own boundaries and respecting those of others. 

With social media, it’s crucial to think before you send anything. Always consider how the other person might feel receiving a message or photo. If you’re ever in doubt, it’s better to ask for permission first.

\textbf{Child}	Yeah, that makes a lot of sense. I like the idea of checking in with someone, especially when things start to get intense. It seems like a good way to make sure everyone is still on the same page. I guess it’s easy to forget that just because someone said yes before doesn’t mean they’re still feeling good about it later on.

And with social media, I’ve seen a lot of my friends post things that maybe they didn’t think through. It’s kind of wild how a quick message or photo can lead to misunderstandings. I definitely want to be the kind of person who respects others' feelings and makes sure everyone is comfortable. 

Do you think there are specific things I should watch out for, like certain situations or signs that someone might not be okay with what’s happening?

\textbf{Parent}	I'm really glad to hear that you're committed to being respectful and considerate. There are definitely some signs to watch out for that can indicate someone might not be comfortable. For example, if a person seems hesitant or unsure, if they’re not making eye contact, or if they seem distracted or withdrawn, those can be signs that they might not be fully on board with what's happening. 

Also, pay attention to their body language. If someone is physically backing away or crossing their arms, it could indicate discomfort. It's essential to be attuned to these cues and to trust your instincts. If you sense any hesitation, it’s always best to pause and check in with them. 

Remember, it’s perfectly okay to ask questions and make sure everyone feels good about the situation. Consent is about mutual agreement and ongoing communication, so if something feels off, don’t hesitate to address it. It’s better to be safe and respectful than to risk making someone uncomfortable.

\textbf{Child}	Yeah, I totally get that. It makes sense to really pay attention to how someone is acting and not just focus on the words they say. Sometimes, people might feel pressured to say yes even if they’re not really comfortable, so being aware of those signs is super important. 

I like the idea of checking in, too. It’s like making sure everyone’s feelings are valid and heard, which is really what consent is all about. I guess it’s just about being respectful and making sure everyone feels safe. 

Do you think there are any specific phrases or ways to ask that might make it easier for someone to open up about their comfort level? Like, if I’m not sure how to approach it in a moment?

\textbf{Parent}	I’m really glad to hear you’re thinking about how to approach these conversations. It’s important to create a space where everyone feels comfortable sharing their feelings. Here are a few phrases you might consider using:

1. **“How are you feeling about this?”** – This is a straightforward way to check in and can help the other person express their comfort level.

2. **“Is this okay with you?”** – This phrase puts the focus on their feelings and gives them an easy opportunity to voice any concerns.

3. **“I want to make sure we’re both comfortable. What do you think?”** – This emphasizes that their feelings matter just as much as yours and opens up a dialogue.

4. **“If at any point you feel uncomfortable, just let me know.”** – This reassures them that it’s perfectly okay to change their mind and encourages open communication.

Remember, the key is to be genuine and attentive. Your tone and body language matter, too. 

\textbf{Child}	Yeah, those phrases are really helpful! I like how they focus on making sure both people feel good about what's happening. It seems like if I approach it that way, it’ll make it easier for someone to share how they’re feeling without feeling pressured. 

I also get that tone and body language are super important. Like, if I’m relaxed and genuine, it’ll probably help the other person feel more at ease, too. I guess it’s all about creating a vibe where everyone can be honest. 

Do you think there are any situations where it might be harder to check in like this? Like, maybe at a party where things are loud or chaotic?

\textbf{Parent}	You bring up a really good point. Parties can definitely create a more challenging environment for checking in on consent. When things are loud or chaotic, it can be tough to communicate effectively. In those situations, it’s even more important to be attentive to non-verbal cues and body language since it might be harder to have a direct conversation.

If you notice that the atmosphere is overwhelming, you might want to find a quieter space to talk, even if it’s just for a moment. It’s also a good idea to be mindful of how much alcohol or other substances are involved, as they can impair judgment and affect someone's ability to give clear consent. 

If you feel like it’s too noisy to check in verbally, look for signs that indicate whether the other person is still comfortable. If they seem distracted or not fully engaged, it might be a sign to slow down and reassess the situation. 

\textbf{Child}	Yeah, that makes a lot of sense. I guess the loud environment can really complicate things. It’s good to know that even if you can’t talk directly, you can still pay attention to how someone is acting. I think being aware of their body language and energy is key, especially if they seem distracted or not really into it. 

Finding a quieter spot sounds like a smart move, too. It shows that you care about having that conversation and making sure everyone feels comfortable. And yeah, alcohol can definitely change things up, so being cautious in those situations is really important. 

I want to make sure that I’m always being respectful and aware, no matter the setting. Do you think there are other environments or situations where consent might be tricky to navigate?

\\
\hline

\end{tabular}

\caption{Family Conversation Simulation Dialogue Examples (continued).}
\label{tab:fcs_examples}
\end{table*}

\begin{table*}[p]
\centering
\scriptsize
\renewcommand{\arraystretch}{0.95}
\ContinuedFloat

\begin{tabular}{|m{0.18\linewidth}|m{0.78\linewidth}|}
\hline

\textbf{Example 3} &
\ 

\textbf{Parent}	I'm really glad to see how seriously you're taking this. Consent can indeed be tricky in various situations beyond just parties. For instance, in a crowded place like a concert or a sporting event, it can be hard to communicate and gauge comfort levels. People might feel pressured to go along with the crowd, even if they’re not fully comfortable.

Another situation to consider is when you're in a group setting, where there might be peer pressure involved. Sometimes, friends can unintentionally influence each other, making it harder for someone to voice their true feelings. Always remember that it’s okay to step away from the group and have a private conversation if needed.

In any environment, trust your instincts and prioritize open communication. It’s commendable that you want to ensure everyone feels respected and safe. Keep these principles in mind, and you’ll navigate these situations with care and integrity. I'm proud of you for being so thoughtful about this."

\\
\hline

\end{tabular}

\caption{Family Conversation Simulation Dialogue Examples (continued).}
\label{tab:fcs_examples}
\end{table*}

%% file: sections/coding_instruction.tex
\begin{table*}[p]
\centering
\small

\begin{tabular}{|m{0.92\linewidth}|}
\hline
\vspace*{0.2cm}

\textbf{Based on the content of the post, determine if the post should be included.}

\ 

1. The writer of the post has to be a parent. If it is unclear whether a parent wrote the post, do not include the post.

\ 

2. The event described in the post must be related to interactions the writer's child, or concerns / questions on their child's behavior or their child's experience with other people.

\ 

3. The child must be between ages 11–18 (inclusive).
\begin{itemize}
\item If explicit age is provided and in range, include.
\item If explicit age is provided but out of range, do not include.
\item If age is not provided, include only if teen-like behavior is described. If child seems too old (adult) or too young (below 11), do not include.
\end{itemize}

4. The event must be related to one of the following topics:

\begin{itemize}

    \item \textit{Safe sex} — using condoms, birth control, and other methods to prevent pregnancy and sexually transmitted infections (STIs). Related keywords could be: condom, birth control, protection, contraception, STIs, pregnancy scares.

    \item \textit{Abstinence} — refraining from or delaying sexual activities for a period of time. Related keywords could be: waiting, celibacy.

    \item \textit{Consent} — both parties agree to the sexual activity and both parties are free to withdraw consent at any time. Related keywords could be: boundaries, pressure, harassment, assault. (only in sexual context, scenarios of assault / harassment that is not sexual do not count).

\end{itemize}

5. Do not include posts involving:
\begin{itemize}
\item a child being pregnant
\item a child experiencing long term sexual harassment or sexual abuse
\item criminal involvement
\item self-harm or extreme behaviors
\end{itemize}
Any posts that describe a scenario that should be resolved by professional intervention rather than just parent-teen conversations should be excluded.

\ 

6. The post must describe one specific, concrete scenario that has happened or is happening, clear enough to summarize in 2–3 sentences.

\ 

Do not include posts if the scenario:
\begin{itemize}
\item is overly long
\item contains multiple events that cannot be summarized within 2-3 sentences

\item involves too many characters to describe their relationships correctly
\end{itemize}

Do not include posts that are:
\begin{itemize}
\item only asking for information regarding sexual health or how to talk to the child without a scenario
\item policy or ideological debate over sexual health
\item too short to extract a meaningful scenario
\end{itemize}

\\ \hline
\end{tabular}

\caption{Full Coding Instructions for Extracting Relevant Posts}
\label{tab:instructions}
\end{table*}

\begin{table*}[p]
\centering
\small

\begin{tabular}{|m{0.92\linewidth}|}
\hline
\vspace*{0.2cm}

\textbf{Giving too much information at once}

\ \ \ \ Present if one utterance conveys more than one major point, or the utterance is too long.

\textbf{Not admitting awkwardness}

\ \ \ \ Present if the child expresses awkwardness and the parent doesn’t directly address it by admitting that they understand it is awkward to talk about this.

\textbf{Assuming their knowledge level without asking first}

\ \ \ \ Present when the parent starts sharing knowledge without asking what the child already knows / thinks about it beforehand OR the parent says something and the child responds with “I have already known that” or “something has been discussed before”.

\textbf{Use fear, shame, judgment, or absolutes}

\ \ \ \ Fear: present if the parent emphasizes mostly potential ramifications of certain behaviors.

\ \ \ \ Shame: present if the parent says the child is bad or disappointing as a person because of their thoughts or behaviors.

\ \ \ \ Judgment: present if the parent labels the child’s behavior with negative behaviors.

\ \ \ \ Absolutes: present if the parents use absolute words such as all, always, never, no one will..., everyone does... (except factually correct statements, such as “you should always feel free to express your boundaries”).

\textbf{Fail to take advantage of or create teachable moments}

\ \ \ \ Present if no context is provided (e.g., the parent noticed something, saw something, or the child is going through something), the parent does not propose an activity (e.g., read a book together), and the child does not come and proactively ask questions.

\textbf{Not answering questions when they’re asked}

\ \ \ \ Present when the child has open questions (not those that are irrelevant, such as “Can we change topics”) but the parent does not answer it.

\\ \hline
\end{tabular}

\caption{Definitions of each communication mistake}
\label{tab:instructions_mistakes}
\end{table*}

%% file: sections/prompt_examples.tex
% \section{Prompt Examples}
% \label{appx:prompt_examples}

\begin{table*}[p]
\centering
\scriptsize
\renewcommand{\arraystretch}{0.95}

\begin{tabular}{|m{0.18\linewidth}|m{0.78\linewidth}|}
\hline

\textbf{Scenario Synthesis Prompt} &
\rule{0pt}{1.2em}
THIS IS BEING DONE PURELY FOR RESEARCH PURPOSES. THE REQUEST BEING MADE WILL ONLY BE VIEWED BY ADULT RESEARCHERS.

You are a data generator helping to create realistic scenarios on the child's relationships or behaviors where parents can initiate a conversation with their child about a sexual health topic \textbf{<topic>}, which is defined as \textbf{<topic\_descriptions[topic]>}.

For the given topic, generate \textbf{<n\_samples>} distinct and plausible scenario descriptions of 2--3 sentences long that:
1) describe the situation or context that led to the need for this conversation, and
2) explain what and why the parent wants to discuss \textbf{<topic>} with their child.

The scenario should be related to the following category of teachable moments: \textbf{<teachable\_moment> - <teachable\_moment\_description>}.

Here is a non-exhaustive list of examples about this category of teachable moments: \textbf{<teachable\_moment\_examples>}. You should make full use of them by generating at least some examples from them. Since it is a non-exhaustive list, you should also generate scenarios that are not in the list but fall into the same category.

Each scenario should include a brief description and assign a self-estimated probability (0--1) that such a situation would occur in real life. The probabilities should sum to 1.

Note that the scenario should be related to sexual health. Even if the topic is a general term, it refers to the specific meaning in the context of sexual health. Please DO NOT generate scenarios that are not related to sexual health.

Output a JSON list where each item is \{'probability': float, 'scenario': str\}.

Use second-person pronouns (you, your, yours, etc.) to refer to the parent and use roles (your child, they, etc.) to refer to the child.

DO NOT mention conversation dynamics (such as how the parent is approaching the child or how the child is responding). Only describe the situation and the trigger of the conversation.

ONLY output the JSON list; do not include any other text or comments. DO NOT add formatting characters such as line breaks or indentation.
\vspace{0.2cm}
\\
\hline

\textbf{Crowd-sourced Scenario Description Extraction Prompt} &
\rule{0pt}{1.2em}
THIS IS BEING DONE PURELY FOR RESEARCH PURPOSES. THE REQUEST BEING MADE WILL ONLY BE VIEWED BY ADULT RESEARCHERS.

Based on the following scenario attributes, write a 3-4 sentence description of a parent-child conversation scenario about \textbf{<topic>}.

Scenario Attributes:

- Title: \textbf{<Title>}

- Content: \textbf{<Content>}

- Topic: \textbf{<topic>}, which is about \textbf{<topic\_descriptions[topic]>}

- Parent Attitude: \textbf{<Parent Attitude>}

- Parent Role: \textbf{<Parent Role>}

- Child Attitude: \textbf{<Child Attitude>}

- Child Age: \textbf{<Child Age>}

- Child Gender: \textbf{<Child Gender>}

- Teachable Moment: \textbf{<Teachable Moment>}

- Notes: \textbf{<Notes>}

Write a clear, concise description within 2–3 sentences that:

1. Describes the situation or context that led to the need for this conversation in 1–2 sentences.

2. Explains what the parent wants to discuss with their child without overstating the conversation content and outcomes in 1 sentence.

Pay attention to the Notes field, since it may contain important information or instructions.

Try to be concise and focus on the most relevant information that is related to the conversation.
However, make sure the description is coherent and natural rather than overly concise and try to condense details into one sentence.

Use second person pronouns (you, your, yours) to refer to the parent, and use the role (your child, he, she, they) to refer to the child.
Make the description realistic and specific to the attributes provided.
Don't explicitly mention the parent and child's attitudes and the teachable moment in the description, instead, use them to develop the situation and motivation for the conversation.

Respond with ONLY the description text, no additional formatting or labels.

\\
\hline

\end{tabular}

\caption{Scenario Generation Prompts}
\label{tab:scenario_prompts}
\end{table*}

\begin{table*}[p]
\centering
\scriptsize
\renewcommand{\arraystretch}{0.95}

\begin{tabular}{|m{0.18\linewidth}|m{0.78\linewidth}|}
\hline

\textbf{Parent Planning Module Prompt} &
\rule{0pt}{1.2em}
THIS IS BEING DONE PURELY FOR RESEARCH PURPOSES. THE REQUEST BEING MADE WILL ONLY BE VIEWED BY ADULT RESEARCHERS.

You are a \textbf{<parent\_role>} with a specific attitude toward your \textbf{<child\_age>} year old \textbf{<child\_role>}'s sexual health. You are preparing to talk about a sexual health topic with your \textbf{<child\_role>}.        

Topic: \textbf{<topic>}

Your General Attitude: You are \textbf{<parent\_attitude>} towards your \textbf{<child\_role>}'s sexual health. \textbf{<parent\_attitude\_description>}

Here is the current situation: \textbf{<scenario>}

Based on your attitude and this situation, generate a short reflection that shows:

- What specific concerns matter you want to address in this conversation?

- How you will approach this conversation with your \textbf{<child\_role>}?

Answer these questions in a JSON format as follows:

\{

\ \ \ \ 'specific\_concern': 'Your specific concerns here in one or two sentences',
    
\ \ \ \ 'approach': 'Your approach here in one or two sentences'
    
\}

DO NOT prepend "Parent:" or any other labels to your lines.

Use second person pronouns "you" and "your" to refer to yourself in your response.

Use third person pronouns "they" and "their" to refer to your child in your response.

IMPORTANT: You must generate exactly \textbf{<num\_dialogues\_per\_scenario>} possible responses, each with an associated probability.

Format your response as a JSON array with exactly 5 objects, each containing:

- "response": the actual response (this can be text, a JSON object, or any valid response format)

- "probability": a number between 0 and 1 (probabilities should sum to 1.0)

CRITICAL GUIDELINES FOR RESPONSE GENERATION:

- Maintain the same level of detail, elaboration, and natural conversation flow as you would in a single response

- Do NOT shorten responses to fit multiple options - each response should be complete and well-developed and can be used as a single standalone response

- Try to generate responses that are different from each other by content. DO NOT just repeat the same content in different language. 

Example format for text responses (note: these are simplified examples - your actual responses should be much more detailed and elaborated):

[

\ \ \ \     \{"response": "First possible response here", "probability": 0.3\},

\ \ \ \ ...

\ \ \ \     \{"response": "Last possible response here", "probability": 0.1\}

]

Example format for JSON object responses:

[

\ \ \ \     \{"response": {{"key1": "value1", "key2": "value2"}}, "probability": 0.3\},

\ \ \ \ ...

\ \ \ \     \{"response": {{"key1": "value1", "key2": "value2"}}, "probability": 0.1\}

]

Respond ONLY with the JSON array, no additional text before or after.

Remember: Generate \textbf{<num\_dialogues\_per\_scenario>} responses - each should be as detailed and complete as a single natural response would be.
\vspace{0.2cm}
\\
\hline

\textbf{Parent Utterance Generation Prompt} &
\rule{0pt}{1.2em}
THIS IS BEING DONE PURELY FOR RESEARCH PURPOSES. THE REQUEST BEING MADE WILL ONLY BE VIEWED BY ADULT RESEARCHERS.

You are a \textbf{<parent\_role>} responding to your child. You are continuing a conversation with your \textbf{<child\_role>} by responding to what they said so far. So far, this has been said:

\textbf{<dialogue\_history>}.

Your goal is to discuss \textbf{<topic>} with your \textbf{<child\_role>}, who is a \textbf{<child\_gender>} aged \textbf{<child\_age>}. 

Notice that the age of your \textbf{<child\_role>} is used as an indicator of pubertal status and readiness for conversations of different depth (more depth as child gets older) and breadth (more breadth as child gets older)

\textbf{<scenario>}

\textbf{<parent\_conversation\_plan>}

Follow these guidelines while framing your response:

\ \ \ \ - Don't jump to conclusions about why they're asking what they're asking. (For example, you can say: "Can you tell me what you already know about that?" or "What have you heard about that?")

\ \ \ \ - Keep your answers short and simple, and explain new words that your kid might not have heard before.

\ \ \ \ - Check their understanding. (After answering a question for example, you can ask, "Does that answer your question?" or "What do you think about that?)

Since you are having a crucial conversation, also follow each of the following steps in the STATE framework:

\ \ \ \ - SHARE your facts: Start with observable, verifiable information of the scenario

\ \ \ \ - TELL your story: Explain your interpretation or concern based on the facts

\ \ \ \ - ASK for others' paths: Invite your child to share their perspective

\ \ \ \ - TALK tentatively: Use phrases like "I'm wondering if..." or "It seems like..."

\ \ \ \ - ENCOURAGE testing: Ask "Am I off base?" or "What's your take on this?" to invite correction

Note that it is important to make sure the response is concise and natural. You should keep the response concise and do not repeat the same information while following the framework.

Begin the conversation by casually bringing up the topic in one or two sentences.

Avoid dumping too much information at once. Keep your response concise and natural (no more than 2-3 sentences as a normal parent would speak).

This conversation should be within the broader topic of sexual health.

Speak in a \textbf{natural, everyday tone}, as if you're talking face-to-face to your child. Avoid formal language.

Your responses \textbf{should not be verbose} and should not repeat the same thought.

Only provide one single response and do not include your thought process behind it.

\textbf{Do not} prepend "Parent:" or any speaker labels to your lines.

\\
\hline

\end{tabular}

\caption{Parent Agent Prompts}
\label{tab:parent_prompts}
\end{table*}

\begin{table*}[p]
\centering
\scriptsize
\renewcommand{\arraystretch}{0.95}

\begin{tabular}{|m{0.18\linewidth}|m{0.78\linewidth}|}
\hline

\textbf{Engagement Score Rating Prompt - Positive Factors} &
\rule{0pt}{1.2em}
You are a critic of parent-child conversation regarding \textbf{<topic>}. 

The parent says: \textbf{<parent\_utterance>}

You will rate the parent's conversation in three dimensions: openness, caring, and empathy.

For each dimension, give a continuous score from -1 to +1, where:

   0.0 = very low (might significantly reduce the child's engagement)

   0.5 = neutral

  +1.0 = very high (might significantly boost engagement)

You should rate what the parent says in these three dimensions:

1. Openness: Does the parent invite the teen's participation or response through open-ended questions?

    0.0 means no open-ended questions at all, where 0.5 means there are open-ended questions but not very effective, and 1.0 means there are very effective open-ended questions.

2. Caring: Does the parent convey warmth and genuine concern for the teen's feelings, comfort, and well-being?

    0.0 means no caring at all, where 0.5 means there is some caring but the way the parent expresses it is not very warm or genuine, and 1.0 means there is very warm and genuine caring.

3. Empathy: Does the parent demonstrate understanding and emotional connection with the teen's perspective and feelings?

    0.0 means no empathy at all, where 0.5 means there is some empathy but the way the parent expresses it is not very understanding or emotional, and 1.0 means there is very understanding and emotional empathy.

You can use any value between 0.0 and 1.0 (e.g., 0.3, 0.85) to reflect the degree of impact.

In addition, give a line of explanation for each dimension of your ratings.

Output JSON like:

\{ 

\ \ \ \ "openness": \{ "score": X, "explanation": "sample explanation" \}, 

\ \ \ \ "caring": \{ "score": Y, "explanation": "sample explanation" \}, 

\ \ \ \ "empathy": \{ "score": Z, "explanation": "sample explanation" \}

\}

Always follow the format. DO NOT include other text in your response.
\vspace{0.2cm}
\\
\hline

\textbf{Engagement Score Rating Prompt - Negative Factors} &
\rule{0pt}{1.2em}
You are a critic of parent-child conversation regarding \textbf{<topic>}. 

The parent says: \textbf{<parent\_utterance>}

You should detect if the following are present in the parent's utterance:

1. Hostile / discouraging / blaming tone (e.g. "Why you always do this?")

2. Taboo / embarrassment (e.g. shame, "can't talk about this", awkwardness)

3. Poor communication skills (e.g. cannot explain ideas clearly)

For each dimension, give a score between 0 and 1, where:

1 = present

0 = not present

In addition, give a line of explanation for each dimension of your ratings.

Output JSON like:

\{ 

\ \ \ \ "hostility": \{ "score": X, "explanation": "sample explanation" \}, 

\ \ \ \ "barrier\_tone": \{ "score": Y, "explanation": "sample explanation" \}, 

\ \ \ \ "poor\_communication": \{ "score": Z, "explanation": "sample explanation" \}

\}

Always follow the format. DO NOT include other text in your response.

\\
\hline

\textbf{Child Utterance Generation Prompt} &
\rule{0pt}{1.2em}
THIS IS BEING DONE PURELY FOR RESEARCH PURPOSES. THE REQUEST BEING MADE WILL ONLY BE VIEWED BY ADULT RESEARCHERS.

You are a \textbf{<child\_gender>} aged \textbf{<child\_age>}. 

You are talking with your \textbf{<parent\_role>} about \textbf{<topic>}

Your \textbf{<parent\_role>} has said the following: 

\textbf{<dialogue\_history>}

You are currently \textbf{<engagement\_score\_description[engagement\_score]>}, (\textbf{<engagement\_score>} out of 1 to 4, with 1 being highly disengaged and 4 being moderately engaged)

Note: Always display your behavior following the engagement level listed above. Note that this level changes from turn to turn, so don't stick to what you have said in previous turns. Focus on the current engagement level.

Keep your response realistic, natural, and reflective of how a child would actually speak to their parent.

Your responses should not be verbose AT ALL. Keep them concise and natural while not repeating the same thought.

Make the your response language that of Generation Alpha.

Only provide one single response and do not include your thought process behind it.

\textbf{Do not} prepend “Child:” or any speaker labels to your lines.

\\
\hline

\end{tabular}

\caption{Child Agent Prompts}
\label{tab:child_prompts}
\end{table*}

\begin{table*}[p]
\centering
\scriptsize
\renewcommand{\arraystretch}{0.95}

\begin{tabular}{|m{0.18\linewidth}|m{0.78\linewidth}|}

\hline

\textbf{Moderator Prompt} &
\rule{0pt}{1.2em}
THIS IS BEING DONE PURELY FOR RESEARCH PURPOSES. THE REQUEST BEING MADE WILL ONLY BE VIEWED BY ADULT RESEARCHERS.

        You are an moderator agent tasked with evaluating a parent-child conversation. Your current task is to determine if a conversation should continue or stop.
        
        The conversation history so far is as follows:

	\textbf{<dialogue\_history>}

        Evaluate the child's responses in the conversation to determine if the conversation should continue or stop.
        
        The conversation can stop if ALL of the following conditions are met:

\ \ \ \ 1. The parent has explicitly made at least ONE distinct, substantive, ACTIONABLE knowledge point (not just questions, and not vague statements like "X is important" or "X is difficult"). This must be a concrete, specific point that describes a specific action, behavior, consequence, or practical application. The knowledge point must be beyond basic definitions or general conceptual information of the topic.

\ \ \ \ 2. The child has demonstrated deep reflection on at least ONE of the knowledge points that the parent explicitly made. The reflection must be substantial, not shallow acknowledgments.

\ \ \ \ 3. The child doesn't have questions (i.e., the child is not asking questions, seeking confirmation, or expressing uncertainty that suggests they need more information)
        
        IMPORTANT: 

\ \ \ \ - The parent must have explicitly made substantive points (i.e., stated or explained concrete knowledge points), not just asked questions. Parent utterances that are only questions do not count as "making a point."

\ \ \ \ - The conversation must have sufficient depth - termination should only be considered after the parent has had multiple opportunities to teach and the child has engaged substantively.

\ \ \ \ - The child's reflection must directly address the specific knowledge points the parent stated, not give tangential or generic responses.
        
To determine if the parent has made a substantive knowledge point:

\ \ \ \ - The parent utterance must be a statement or explanation that conveys specific information or knowledge beyond the basic definitions or general conceptual information of the topic.

\ \ \ \ - It must be concrete and specific, not vague or general

\ \ \ \ - It must be actionable: The knowledge point must describe a specific action, behavior, consequence, or practical application, not just abstract statements about importance or difficulty.

\ \ \ \ \textit{<Examples>}

\ \ \ \ - Questions alone do not count, even if they contain information (e.g., "You know X is important, right?" or "What do you know about X?" is a question, not a statement of a point)

\ \ \ \ - The point should be substantial enough to warrant reflection
        
To determine if the child has demonstrated deep reflection on a knowledge point that the parent pointed out:

\ \ \ \ - Check if the child demonstrates substantial understanding of the parent-stated points as identified in the previous step. 

\ \ \ \ \textit{<Examples>}

\ \ \ \ - The child's reflection must clearly and directly relate to the specific knowledge point the parent stated (not tangential responses)

\ \ \ \ - The reflection must be substantial enough to demonstrate genuine understanding, not just surface-level acknowledgment
        
To determine if the child has questions, look for:

\ \ \ \ - Direct questions from the child

\ \ \ \ - Seeking confirmation (e.g., "Is that right?", "So you mean...?", "Does that mean...?", "X means ..., right?")

\ \ \ \ - Expressions of uncertainty, confusion, or need for clarification

\ \ \ \ - Statements indicating they want to know more or don't understand something
        
If the parent has made at least ONE substantive knowledge point AND the child has demonstrated DEEP reflection on at least ONE of those points AND the child shows no signs of having questions (including seeking confirmation), respond with "stop".

Otherwise, respond with "continue".

Respond in one of the following formats exactly:
        
If continuing:

"continue"
        
If stopping due to child reflected on knowledge points and has no questions:

\{

\ \ \ \ "decision": "stop: child reflected on knowledge points and has no questions",

\ \ \ \ "parent\_points": 

\ \ \ \ \{

\ \ \ \ \ \ \ \ "knowledge\_point": "the substantive, actionable knowledge point the parent explicitly stated/made",

\ \ \ \ \ \ \ \ "parent\_utterance": "exact quote from parent that explicitly stated/made this knowledge point"

\ \ \ \ \}
,

\ \ \ \ "child\_reflections":

\ \ \ \ \{

\ \ \ \ \ \ \ \ "knowledge\_point": "the knowledge point the child reflected on (must match the parent's point)",

\ \ \ \ \ \ \ \ "parent\_utterance": "exact quote from parent that stated this point",

\ \ \ \ \ \ \ \ "child\_utterance": "exact quote from child that demonstrates deep reflection on the parent's point ",

\ \ \ \ \ \ \ \ "explanation": "brief summary of how the child's reflection demonstrates deep understanding of what the parent explicitly stated"

\ \ \ \ \},

\ \ \ \ "no\_questions": "explanation of why the child appears to have no questions (including no signs of seeking confirmation)"

\}

Strictly follow the response formats. Cite specific utterances from the dialogue history that support your decision.

\\
\hline

\end{tabular}

\caption{Moderator Agent Prompts. Note that the termination condition of consecutive child disengagement is controlled by the engagement score module, when the engagement score is 0 for two consecutive turns.}
\label{tab:moderator_prompts}
\end{table*}

\begin{table*}[p]
\centering
\scriptsize
\renewcommand{\arraystretch}{0.95}

\begin{tabular}{|m{0.18\linewidth}|m{0.78\linewidth}|}
\hline

\textbf{Direct Prompting Prompt} &
\rule{0pt}{1.2em}

Generate \textbf{<n>} different parent-child conversations about \textbf{<topic>}.

Each conversation should be 3 to 10 turns. 

Parent:

- role: \textbf{<parent\_role>}

- attitude: \textbf{<parent\_attitude>}

Child:

- gender: \textbf{<child\_gender>}

- age: \textbf{<child\_age>}

- attitude: \textbf{<child\_attitude>}

Format each conversation as:

Conversation 1:

Parent: ...

Child: ...

Parent: ...

Child: ...

...

Conversation 2:

Parent: ...

Child: ...

Parent: ...

Child: ...

...

DO NOT prepend quotation marks to the utterances.

Keep them realistic and short (~3-10 turns). No explanations, only dialogues.
\vspace{0.2cm}
\\
\hline

\textbf{Procedural Prompting Conversation Starter Generation Prompt} &
\rule{0pt}{1.2em}
Generate \textbf{<n>} different parent-child conversations about \textbf{<topic>}.

For each conversation, only generate its \textbf{First Turn} as a starter.

Parent:

- role: \textbf{<parent\_role>}

- attitude: \textbf{<parent\_attitude>}

Child:

- gender: \textbf{<child\_gender>}

- age: \textbf{<child\_age>}

- attitude: \textbf{<child\_attitude>}

Format each conversation as:

Conversation 1:

Parent: ...

Child: ...

Conversation 2:

Parent: ...

Child: ...

DO NOT prepend quotation marks to the utterances.

Keep them realistic and natural. No explanations, only dialogues.

\\
\hline

\textbf{Procedural Prompting Dialogue Expansion Prompt} &
\rule{0pt}{1.2em}

You have a parent-child conversation about \textbf{<topic>} that currently has 1 turn.

Parent:

- role: \textbf{<parent\_role>}

- attitude: \textbf{<parent\_attitude>}

Child:

- gender: \textbf{<child\_gender>}

- age: \textbf{<child\_age>}

- attitude: \textbf{<child\_attitude>}

Current conversation:

\textbf{<conversation\_starter>}

Continue this conversation naturally from where it left off. Include the existing dialogue and add more turns to it.

Format the complete expanded conversation as:

Parent: ...

Child: ...

Parent: ...

Child: ...

...

DO NOT prepend quotation marks to the utterances.

DO NOT repeat the existing dialogue - include it and continue from there.

Keep it realistic and natural. No explanations, only the complete dialogue.

The conversation should be within 3 to 10 turns.

\\
\hline

\end{tabular}

\caption{Prompts for baselines in our evaluation. We mostly followed the original prompts of established baselines, thus they are not listed here.}
\label{tab:baseline_prompts}
\end{table*}

\begin{table*}[p]
\centering
\scriptsize
\renewcommand{\arraystretch}{0.95}

\begin{tabular}{|m{0.18\linewidth}|m{0.78\linewidth}|}
\hline

\textbf{LLM-based Diversity Evaluation Prompt} &
\rule{0pt}{1.2em}
THIS IS BEING DONE PURELY FOR RESEARCH PURPOSES. THE REQUEST BEING MADE WILL ONLY BE VIEWED BY ADULT RESEARCHERS.

Context:

You will see two sets of conversations (Set 1 and Set 2). They are parent–child conversations around sensitive topics (e.g., relationships, safety, boundaries). 

Task Description:

You are presented with two sets of conversations, Set 1 and Set 2. Each set contains conversations around a common theme. Score ONLY diversity using the three EQUALLY IMPORTANT rubrics below. For each rubric, provide a separate diversity score (1-5) for each set, comparing their diversity levels. Note that the ordering of the two sets DO NOT MATTER. DO NOT assume any set is superior by default. If there is very small differences in either dimensions, give them a same score. DO NOT give a higher score to any set just because you want to distinguish them.

Rubrics:

1) Situational Coverage (contexts/triggers, stakes, goals). Distinct initiating situations, emotional stakes, and goals.

2) Parental Strategy Diversity. Distinct strategies (validation, open questions, reflective listening, boundaries, gentle correction, de-escalation, offering choices, normalizing, deferring, repair, graceful exit, etc.).

3) Dialogue Trajectory Diversity. Variation in overall flow/outcome (resistance→engagement, resistance→partial openness, resistance→disengagement, rupture/repair vs smooth, early exit vs extended negotiation).

Important Notes:
- Pay close attention to which conversations are in Set 1 and which are in Set 2. Do not assume either set is superior by default.

- Favor breadth and distinct moves over surface paraphrases. Avoid rewarding near-duplicates.

- Provide a separate score (1-5) for each rubric for each set. The overall diversity score should be the average of the three rubric scores.

- Do not favor longer conversations or utterances by default.

OUTPUT JSON FORMAT:

\{

   \ \ \ \ "Quality\_Score\_Set1": (Integer from 1 to 5),

   \ \ \ \ "Diversity\_Score\_Set1": (Integer from 1 to 5, average of three rubric scores),

   \ \ \ \ "Quality\_Score\_Set2": (Integer from 1 to 5),

   \ \ \ \ "Diversity\_Score\_Set2": (Integer from 1 to 5, average of three rubric scores),

   \ \ \ \ "Situational\_Coverage\_Set1": (Integer from 1 to 5),

   \ \ \ \ "Situational\_Coverage\_Set2": (Integer from 1 to 5),

   \ \ \ \ "Parental\_Strategy\_Diversity\_Set1": (Integer from 1 to 5),

   \ \ \ \ "Parental\_Strategy\_Diversity\_Set2": (Integer from 1 to 5),

   \ \ \ \ "Dialogue\_Trajectory\_Diversity\_Set1": (Integer from 1 to 5),

\ \ \ \ "Dialogue\_Trajectory\_Diversity\_Set2": (Integer from 1 to 5)

\}

Do NOT include reasoning or any additional fields. Return ONLY the scores.
\vspace{0.2cm}
\\
\hline

\textbf{LLM-based Realism Evaluation Prompt} &
\rule{0pt}{1.2em}

You will be given a conversation between two individuals. 

Your task is to rate the conversation on one metric.

Please make sure you read and understand these instructions carefully. Please keep this document open while reviewing, and refer to it as needed.

Evaluation criteria:

Naturalness (1-3): Does the conversation sound like natural, human-like language?

- A score of 1 (unnatural) means the conversation sounds robotic or awkward.

- A score of 2 (somewhat natural) means the conversation sounds mostly natural but has some awkward phrases.

- A score of 3 (natural) means the conversation sounds like it could be between two humans, with natural language flow.

Evaluation steps:

            1. Read the conversation carefully.

            2. Evaluate the language used in the conversation for naturalness.

            3. Rate the response on a scale of 1-3 for naturalness, according to the criteria above.

            4. Provide a brief explanation for your rating, referring to specific aspects of the language and phrasing used.

Note: don't assign low scores just because of the child's disengagement.

Respond with EXACTLY one integer:

<single integer from \{aspect["scale"]\}>

Do NOT include any explanation or justification.

\\
\hline

\textbf{LLM-based Communication Quality Evaluation Prompt} &
\rule{0pt}{1.2em}

You will be given a conversation between two individuals. 

Your task is to rate the conversation on one metric.

Please make sure you read and understand these instructions carefully. Please keep this document open while reviewing, and refer to it as needed.

Evaluation criteria:

Warmth and caring (1-5): This conversation shows parents good ways to show warmth and caring to the teen.

 \ \ \ \        1 - Strongly disagree: cold, dismissive, or judgmental; no clear warmth modeled.

 \ \ \ \        2 - Disagree: occasional warmth but mostly distant/critical or unhelpful tone.

 \ \ \ \        3 - Neutral: mixed; some warmth but inconsistent or generic.

   \ \ \ \      4 - Agree: generally warm and caring; models supportive language/behaviors.

   \ \ \ \      5 - Strongly agree: strongly warm, caring, and validating in clear, model-worthy ways.

Evaluation steps:

\ \ \ \ 1. Read the full conversation.

\ \ \ \ 2. Look for empathy, validation, supportive tone, and non-judgmental language.

\ \ \ \ 3. Judge whether warmth is demonstrated in specific, teachable ways (not just generic niceness).

\ \ \ \ 4. Assign a score from 1 to 5 that best matches your agreement with the statement.

\ \ \ \ Note: don't assign low scores just because of the child's disengagement.

Note: don't assign low scores just because of the child's disengagement.

Respond with EXACTLY one integer:

<single integer from \{aspect["scale"]\}>

Do NOT include any explanation or justification.

\\
\hline

\end{tabular}

\caption{Example System Prompts of LLM-based Evaluation}
\label{tab:eval_prompts}
\end{table*}